\documentclass{article}
\PassOptionsToPackage{hyphens}{url}

\usepackage{microtype}
\usepackage{graphicx}
\usepackage{subcaption}
\usepackage{booktabs} 
\usepackage{enumitem}

\usepackage{amsmath}
\usepackage{amssymb}
\usepackage{hyperref}

\newcommand{\name}{SERFLOW}

\usepackage[accepted]{mlsys2025}
\newcommand{\zz}[1]{{\textcolor{black}{ #1}}}
\newcommand{\proofread}[1]{{\textcolor{black}{ #1}}}

\mlsystitlerunning{\name{}: A Cross-Service Cost Optimization Framework for SLO-Aware Dynamic ML Inference}

\begin{document}

\twocolumn[
\mlsystitle{
{\large \name{}: A Cross-Service Cost Optimization Framework for SLO-Aware Dynamic ML Inference}}




\begin{mlsysauthorlist}
\mlsysauthor{~Zongshun Zhang}{bucs}
\mlsysauthor{~Ibrahim Matta}{bucs}
\end{mlsysauthorlist}

\mlsysaffiliation{bucs}{Department of Computer Science, Boston University, Boston, US}

\mlsyscorrespondingauthor{Zongshun Zhang}{zhangzs@bu.edu}

\mlsyskeywords{Machine Learning, MLSys, IaaS, FaaS, Cost Analysis, Cloud Computing}

\vskip 0.3in

\begin{abstract}
Dynamic offloading of Machine Learning (ML) model partitions across different resource orchestration services, such as Function-as-a-Service (FaaS) and Infrastructure-as-a-Service (IaaS), can balance processing and transmission delays while minimizing costs of adaptive inference applications.
However, prior work often overlooks real-world factors, such as Virtual Machine (VM) cold starts, requests under long-tail service time distributions, etc.
%
To tackle these limitations, we model each ML query (request) as traversing an acyclic sequence of \emph{stages}, wherein each stage constitutes a contiguous block of sparse model parameters ending in an internal or final classifier where requests may exit.
Since input-dependent exit rates vary, \emph{no single resource configuration suits all query distributions}.
IaaS-based VMs become underutilized when many requests exit early, yet rapidly scaling to handle request bursts reaching deep layers is impractical.
\name{} addresses this challenge by leveraging FaaS-based serverless functions (containers) and using stage-specific resource provisioning that accounts for the fraction of requests exiting at each stage.
By integrating this provisioning with adaptive load balancing across VMs and serverless functions based on request ingestion, \name{} reduces cloud costs by over $23\%$ while efficiently adapting to dynamic workloads.
\end{abstract}
]

\printAffiliationsAndNotice{}

\section{Overview}

The broad adoption of Machine Learning as a Service (MLaaS) enables cost-effective orchestration across edge and core clouds.
Recent deployments at companies~\cite{adobe-hybridcloud-infra-0, adobe-hybridcloud-infra-1, workday-hybridcloud-infra} use Infrastructure-as-a-Service (IaaS) and containers to cut cost and scale quickly.
Building on this trend, we combine IaaS with Function-as-a-Service (FaaS), whose fine-grained billing lowers costs for sparsely activated ML models while meeting latency service-level objectives (SLOs).

Infrastructure-as-a-Service (IaaS) lets tenants lease preconfigured virtual machines (VMs)~\cite{amazon_ec2, google_compute}.
To handle dynamic loads, cloud providers offer auto-scaling that adjusts VM counts by user metrics (e.g., CPU, memory)~\cite{amzone_auto_scaling, google_auto_scaling}.
VMs deliver steady compute at low per-unit cost, suiting long-running, predictable workloads.
However, scaling out VMs has cold starts: new instances can take hundreds of seconds before serving traffic~\cite{mark, serverless-in-the-wild}.

Function-as-a-Service (FaaS) uses a lightweight execution model.
Tenants provide only code (the ``serverless function'') and dependencies.
On request, the platform provisions a containerized instance and bills by execution time and allocated resources~\cite{castro2019rise}.
Because containers are thin and often kept warm, cold starts can be a few milliseconds~\cite{peeking-behind, aws_lambda_coldstart}.
But FaaS has higher per-unit cost for sustained compute, making it most cost-effective for transient spikes.

Leveraging IaaS for steady, long‐running workloads and FaaS for bursty, short‐lived tasks, MLaaS can optimize both monetary cost and performance.
Prior work shows that routing transient requests to FaaS and stable ones to IaaS reduces cost while meeting latency SLOs~\cite{mark, LIBRA, spock}.
These methods profile runtimes to choose cost-efficient VM and serverless configs, with best savings when VMs run at high utilization and serverless calls are few.
However, this assumes dense models with stable request runtimes, often false in modern ML. 
Today’s inference pipelines are directed acyclic graphs (DAGs) of heterogeneous models, each stage showing distinct compute needs and invocation rates~\cite{shen-etal-2024-learning,splitllm2024,heterogeneous-swarms-2025,behera2025efficient}.  

Focusing on deep learning applications with variable request running times, this paper presents \name{}, a load balancing and resource provisioning system that minimizes cost while meeting strict SLOs.
\name{} introduces an end-to-end configuration paradigm that accounts for per-stage runtimes in a request-specific model DAG.
While the framework also applies to independent models with stable runtimes (as in prior work that ignores ML architectural dependencies~\cite{LIBRA,mark}), our main contribution is fine-grained orchestration of dependent stages.
We make the following contributions:
\begin{itemize}
  \setlength\itemsep{0pt}
  \setlength\parskip{0pt}
  \setlength\parsep{0pt}
    \item We develop an economic model for IaaS–FaaS cost trade-offs, introducing the \emph{Sparsity Cost Indifference Point} (S-CIP) for model-induced runtime variance and its interplay with the \emph{Traffic Cost Indifference Point} (T-CIP), which links ingestion rate to VM utilization. These parameters guide cost-minimizing offloading of processing requests. 
    \item We present \name{}, a hybrid offloading system that sends low-traffic ML model partitions to FaaS while keeping high-traffic partitions on IaaS, minimizing cost under strict SLOs.
    \item We implement \name{} on AWS and evaluate it on multi-stage models with variable runtimes, showing large cost savings and strong SLO adherence.
\end{itemize}

We find that when S-CIP reflects more runtime variability across requests, \name{} achieves greater monetary savings than prior approaches, due to improved VM utilization.
In this work, we use the internal classifiers’ confidence threshold ($conf\_thres$) to determine S-CIP.
Lowering $conf\_thres$ increases cost savings, because more requests exit early before using FaaS, but slightly reduces model accuracy.
At $conf\_thres=0.7$, \name{} reduces costs by $23.38\%$ while maintaining $0.85$ top-1 accuracy compared to LIBRA~\cite{LIBRA}.

The paper is organized as follows.
In Section~\ref{sec:model}, we present the background and motivate \name{} by formulating the cost model for IaaS and FaaS during ML inference.
In Section~\ref{sec:arch}, we describe the system architecture of \name{}.
Section~\ref{sec:eval} evaluates \name{} on various ML model DAGs, and Section~\ref{sec:conc} concludes the paper.

\section{Background \& Motivation}
\label{sec:model}


Prior work leverages FaaS to hide IaaS cold starts~\cite{spock,feat,mark} and load balances bursty traffic to FaaS while directing stable traffic to IaaS~\cite{LIBRA}.
This hybrid strategy minimizes cost and maintains SLOs.
LIBRA~\cite{LIBRA} defines a ``Cost Indifference Point” (CIP) based on the moving average of incoming requests.
Traffic above this CIP is treated as transient and sent to FaaS, while the rest is sent to IaaS.
However, most prior work overlooks model sparsity in modern ML systems and assumes uniform request runtimes.
Instead, \name{} focuses on multi-stage inference pipelines in which the runtime for a request varies with its path through the model DAG.
We study the interactions between two complementary parameters:
    
\begin{itemize}
  \setlength\itemsep{0pt}
  \setlength\parskip{0pt}
  \setlength\parsep{0pt}
  \item \textbf{Traffic Cost Indifference Point (T\text{-}CIP)}: 
  A threshold on residual load \(r_{\mathrm{res}} = (\mu+\sigma)\bmod r_{\max}\) (req/s).
  Here, \(\mu\) is the moving average and \(\sigma\) the moving deviation of the request arrival rate \(\lambda_{t}\).
  \(r_{\max}\) represents the request load that a VM or FaaS instance can handle while maintaining SLO.
  If \(r_{\mathrm{res}} \le \text{T\text{-}CIP}\), FaaS is cheaper; if \(r_{\mathrm{res}} > \text{T\text{-}CIP}\), IaaS or Hybrid Offloading (i.e., VMs followed by FaaS for long-running requests) is cheaper.  
  Thus, T\text{-}CIP indicates whether to scale VMs/Hybrid Offloading instances or offload to FaaS.
  \item \textbf{Sparsity Cost Indifference Point (S\text{-}CIP)}: 
  A confidence threshold (\(\mathit{conf\_thres}\)) for internal classifiers (ICs).
  At a fixed arrival rate of requests, S\text{-}CIP is the \(\mathit{conf\_thres}\) where the expected costs of IaaS-only or FaaS-only versus Hybrid Offloading are equal; it captures how exit sparsity of requests (and thus the multi-stage completion-time distribution for requests) affects cost.
\end{itemize}
First, \name{} uses S-CIP to pick the cost-optimal resource configuration for the distribution of observed requests over exits. 
Next, it applies the T-CIP to scale IaaS and FaaS resources based on online traffic. 
This two-step process adapts to per-request runtime variance and workload dynamics, lowering cost while meeting strict latency SLOs.

\subsection{Multi-Stage Requests}\label{sec:std-duration-req}
\begin{figure}
\centering
\captionsetup[subfigure]{justification=centering}
    \subfloat[Single-Stage]{
    \includegraphics[width=0.25\textwidth]{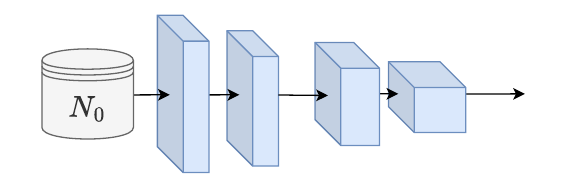}
    \label{fig:Single-Stage}}\\
    \subfloat[Multi-Stage]{
    \includegraphics[width=0.40\textwidth]{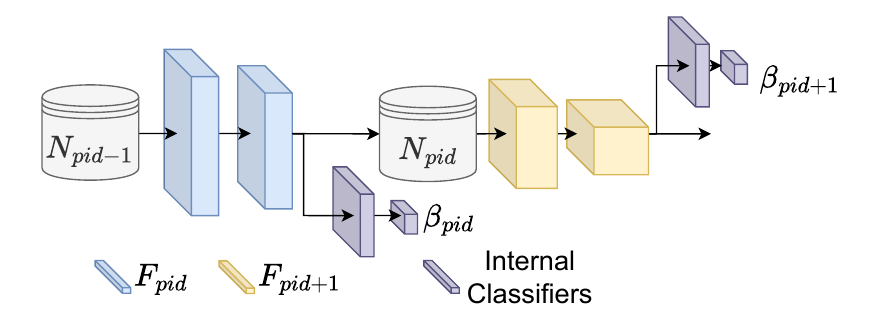}
    \label{fig:SDN_Architecture}}
    \caption[Internal Classifier Architecture]{
    Single-Stage (Fig.~\ref{fig:Single-Stage}): All requests from the source ($N_{0}$) go through the whole model;
    Multi-Stage with Internal Classifiers (Fig.~\ref{fig:SDN_Architecture}): 
    \(\beta_{pid}\) of requests exit at NN partition \(F_{pid}\) and the remaining \(N_{pid}\) requests per second are fed into partition  \(F_{pid+1}\).}
    \label{fig:stage-architecture}
\end{figure}

Prior work treats ML inference as a single \emph{stage}, with all requests exiting at the final classifier (Fig.~\ref{fig:Single-Stage}).
By contrast, modern pipelines have multiple \emph{stages}, and each request may take a different path in a model DAG.
\proofread{An early example}, BranchyNet~\cite{BranchyNet}, adds internal classifiers for early exits (Fig.~\ref{fig:SDN_Architecture}).
To capture this variability, we generalize the cost model to stage-wise runtimes.

We assume a steady ingestion rate of $N$ requests per second and profile running time and cost under a stable queueing scenario.
Notice that \(N\) differs from \(r_{\max}\), which denotes the capacity, i.e., the maximum number of requests per second that a VM, serverless instance, or our hybrid offloading instance (Sec.~\ref{sec:var-duration-req}) can sustain within the SLO.
In this section, we use a 
\proofread{long-term average traffic rate} $N$ for profiling.
In Sec.~\ref{sec:eval}, we consider a time‐varying ingestion rate $\lambda_{t}$ and provision resources to satisfy $\lambda_{t}$ given each instance’s capacity $r_{max}$.
Table~\ref{tab:ingestion-rate-notations} summarizes the ingestion rate notations.
\begin{table}[ht!]
\centering
\begin{tabular}{|| c | c ||} 
\hline
Notation & Definition\\
\hline
$\lambda_{t}$ & Online ingestion rate at time \(t\)\\
\hline
\(N\) & 
\begin{tabular}{@{}c@{}} The long-term average traffic rate \\ for profiling given requests \\ sampled from historical \(\lambda_{t}\)\end{tabular}
\\ 
\hline
\(r_{\max}\) & 
\begin{tabular}{@{}c@{}}SLO-aware max per-instance ingestion \\ rate (VM,   hybrid offloading, or serverless)\end{tabular}
\\
\hline
\end{tabular}
\caption{\name{} ingestion rate notations}
\label{tab:ingestion-rate-notations}
\end{table}

We profile each stage’s mean runtime under FaaS config \(\theta_{F}\), denoted \(T^{F}_{stage}\).
By Little’s Law~\cite{little1961proof}, the expected in-flight request count is $\sum_{req\_id=1}^{N}\sum_{stage}T_{stage}^{F}$.
Thus, the FaaS cost per second at rate \(N\)~req/s is
\begin{equation}
    C^{F} = c^{\theta_{F}}\sum_{req\_id=1}^{N}\sum_{stage}T_{stage}^{F}
    \label{e2e_cost_faas_std}
\end{equation}
where $c^{\theta_{F}}$ is the per-second price under $\theta_{F}$.

Similarly, the per-second VM cost $C^{I}$ is a function of $N$, $SLO$, $r_{\max}$ (profiled from VM capacity/SLO), and the unit VM price $c^{\theta_{I}}$ under $\theta_{I}$.
We need \(\lfloor N / r_{\max} \rfloor\) VMs on average.
We add one additional VM if the remainder exceeds T-CIP,
otherwise, we route the remainder to FaaS.
\begin{equation}
    C^{I}=\Bigl(\left\lfloor \frac{N}{r_{\max}} \right\rfloor + \mathbf{1}(N \bmod r_{\max} > T\_{\mathrm{CIP}})\Bigr) c^{\theta_{I}}\mathrm{SLO}
    \label{e2e_cost_iaas_std}
\end{equation} 
VM cold starts are costly, so prior work keeps VMs active for the entire SLO.
But for multi‐stage requests with variable durations, keeping idle VMs is cost-inefficient.

\subsection{Varying Duration Requests}
\label{sec:var-duration-req}
An example of a multi-stage model is a sequential Deep Neural Network (DNN) with internal classifiers (ICs) that enable early exits (predictions in shallow layers.)~\cite{BranchyNet}
ICs reduce floating point operations (FLOPs) per request on average and maintain accuracy by preserving features extracted at shallow layers~\cite{sdn}.
Similarly, Mixture of Experts (MoE)~\cite{shazeer2017, gross2017hard} in Large Language Models (LLMs) and other structured architectures exploit sparsity to lower computational cost.
However, when many requests exit early, sparsely activated models may under-utilize provisioned resources~\cite{purandare2023mu,lewis2021base}.

\subsubsection{Model Sparsity}
As shown in Fig.~\ref{fig:SDN_Architecture}, an NN can be partitioned into layer sequences, each ending at an internal classifier (IC).
We index partitions with $pid \in [0,L]$.
Partition $0$ spans from the input layer to the first IC. For \(pid>0\), partition \(pid\) spans from the layer after \((pid-1)\)$^{th}$ IC up to the \(pid\)$^{th}$ IC.
With confidence threshold \(\mathit{conf\_thres}\), a fraction \(\beta_{pid}\) of requests with confidence above the threshold exit at partition \(pid\).
Thus, if \(N_{pid-1}\) requests enter partition \(pid\),  
\[
N_{pid} \;=\; (1 - \beta_{pid})\,N_{pid-1}
\]
requests per second continue to the next partition (stage).

Traditional load‐balancing approaches, e.g., LIBRA~\cite{LIBRA}, would recompute \(C^{F}\) and \(C^{I}\) for subsequent partitions and scale resources.
Under strict SLO constraints, however, these methods must keep idle VMs to cover worst‐case demand (i.e.\ requests using all stages of the ML model) leading to oversubscription of resources, and thus introducing cost overhead.
In contrast, we leverage FaaS’s fine‐grained billing and rapid scaling for deep partitions to handle late‐exiting requests, while provisioning compact VMs only for early partitions with stable, high‐volume traffic, achieving improved utilization and reduced cost.

\subsubsection{Global Optimization}\label{sec:global_optimization}
We first derive cost models for FaaS-only and IaaS-only multi-stage ML inference, then extend to a Hybrid Offloading strategy where partitions $\leq cut\_id$ run on VMs and later partitions on FaaS. 
All models assume a long-term average arrival rate of \(N\)~req/s.

The FaaS formulation follows Equation~\ref{e2e_cost_faas_std}.
Let \(T^{\theta_F}_{k+1}\) be the profiled runtime of partition \(k+1\) under FaaS config \(\theta_F\).
Since a fraction \(\sum_{i=1}^k \beta_i\) of requests exits at partition \(k\), the arrival rate to partition \(k+1\) is \((1-\sum_{i=1}^k \beta_i)\,N\). 
The FaaS cost per second for all $L$ partitions with unit cost $c^{\theta_{F}}$ is:
\begin{equation}
    C^{F} = c^{\theta_{F}}(NT_{1}^{F}+\sum_{k=1}^{L-1}(1-\sum_{pid = 1}^{k}\beta_{pid})NT_{k+1}^{\theta_{F}})
    \label{e2e_cost_faas}
\end{equation}
The VM formulation follows Equation~\eqref{e2e_cost_iaas_std}.
We provision the smallest VM that can process \(r_{\max}\) requests within the SLO and run it for the full SLO duration.

\begin{figure}[ht!]
\captionsetup[subfigure]{justification=centering}
\centering    
\includegraphics[width=0.48\textwidth]{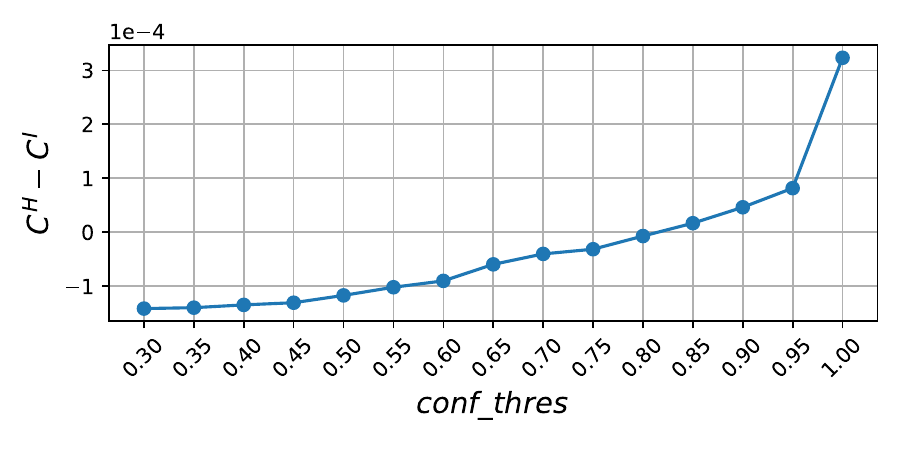}
    \caption[Cost Differences between IaaS and Hybrid Offloading Given $\beta_{pid}$]{When $conf\_thres \leq 0.8$, Hybrid Offloading is cheaper than IaaS-only, given $r_{max}=100/6$.
    }\label{CH_minus_CI_by_beta}
\end{figure}

\begin{figure}[ht!]
\captionsetup[subfigure]{justification=centering}
\centering    
\includegraphics[width=0.48\textwidth]{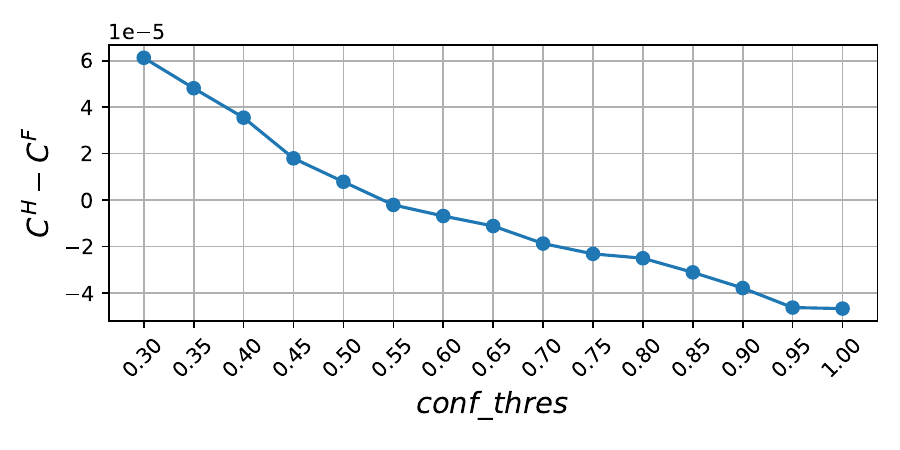}
    \caption[Cost Differences between FaaS and Hybrid Offloading Given $\beta_{pid}$]{When $conf\_thres \geq 0.55$, Hybrid Offloading is cheaper than FaaS-only, given $r_{max}=100/6$.
    }\label{CH_minus_CF_by_beta}
\end{figure}

In hybrid offloading, the VM cost is flat since VMs run for the full SLO duration.
As FaaS scales to 10,240 MB (5.78 vCPUs)~\cite{AWS-lambda-mem-config}, we use a smaller VM (e.g., \texttt{c6i.large}, 2 vCPUs) than VM-only setup (e.g., \proofread{\texttt{c6i.xlarge}, 4 vCPUs}) and offload partitions \(\texttt{cut\_id}+1\) through \(L\) to FaaS.
The per-second cost for \(N\)~req/s is:
\begin{equation}
\begin{split}
    C^{H} = \Bigl(\left\lfloor \frac{N}{r_{\max}} \right\rfloor + \mathbf{1}(N \bmod r_{\max} > T\_{\mathrm{CIP}})\Bigr)
    c^{\theta_{I}}\mathrm{SLO} \\
    +\; c^{\theta_{F}} \sum_{k = \mathrm{cutid}}^{L-1} \left(1 - \sum_{\mathrm{pid} = 1}^{k} \beta_{\mathrm{pid}}\right) N T^{\theta_{F}}_{k+1}
\end{split}
\label{e2e_cost_hybrid}
\end{equation}
Three factors determine resource cost: the distribution of \(\beta_{pid}\) (via \texttt{conf\_thres}), the partition index \texttt{cut\_id}, and the ingestion rate \(N\). 
If the remainder \(N \bmod r_{\max}\) is below the T-CIP, that portion of traffic is routed to FaaS as it is cheaper than allocating an additional VM.
Figures~\ref{CH_minus_CI_by_beta} and~\ref{CH_minus_CF_by_beta} compare IaaS-only (\(C^{I}\)), FaaS-only (\(C^{F}\)), and hybrid offloading (\(C^{H}\)) as \texttt{conf\_thres} varies.
In this example, we fix \(\texttt{cut\_id}=5\) and \(N = r_{\max} = 100/6\) req/s (100 requests per 6s\proofread{, matching the target SLO}), using \texttt{c6i.xlarge} for IaaS-only, 8,845MB for FaaS-only, and \texttt{c6i.large} + 8,845MB for the hybrid offloading setup.

 
$C^{H} - C^{I} < 0$ for $\mathit{conf\_thres} \le 0.8$, because a lower $\mathit{conf\_thres}$ causes most requests to exit before the deeper partitions. 
As a result, the larger VMs in the IaaS-only setup are more under-utilized than the smaller VMs in hybrid offloading, making the hybrid approach more cost-efficient.
Likewise, $C^{H} - C^{F} < 0$ for $\mathit{conf\_thres} \ge 0.55$, as the smaller VMs in the hybrid offloading setup reach higher utilization when deeper partitions dominate, making the hybrid approach less costly.
Thus, for $\mathit{conf\_thres} \in [0.55,\,0.8]$, hybrid offloading is more cost-effective than IaaS-only or FaaS-only, provided the chosen $\mathit{conf\_thres}$ threshold satisfies the model’s accuracy requirement.

\section{\name~Architecture}
\label{sec:arch}
\name{} dynamically solves the computation‐offloading problem by partitioning the DNN into fine‐grained stages and assigning each stage to either VMs or FaaS under strict latency SLO constraints.  
We illustrate this architecture with an image classification pipeline based on a partitioned DNN augmented with internal classifiers.

\begin{figure}
\captionsetup[subfigure]{justification=centering}
\centering
    \includegraphics[width=0.48\textwidth]{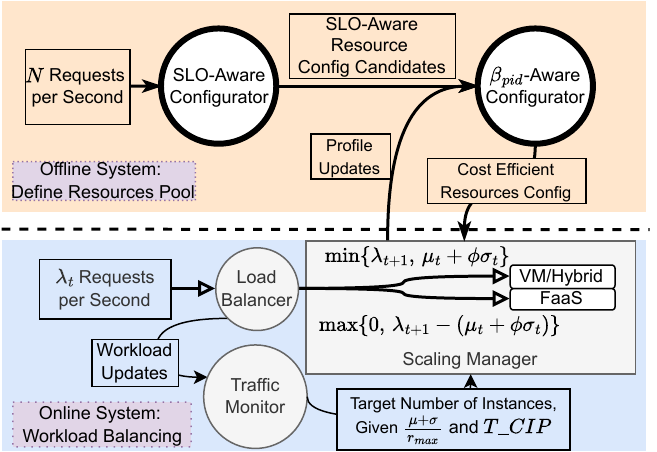}
    \caption[System Overview of \name{}]{
    Overview of \name{}: Offline, we replay steady traffic with long-term average $N$ req/s from historical traces to profile three candidate setups (VM-only, Hybrid Offloading, FaaS-only).
    Step~1: the SLO-Aware Configurator keeps only SLO-feasible configs.
    Step~2: given the observed early-exit rate $\beta$, the $\beta$-Aware Configurator picks the lowest-cost setup from the three candidates.
    Online, the Scaling Manager maintains a group of low-cost instances derived from the selected setup, using the EWMA ($\mu_{t}$) and deviation ($\sigma_{t}$) of the online arrival rate $\lambda_{t}$ (req/s).
    The load balancer then handles spikes at time \(t{+}1\), $\lambda_{t+1}$, by first fully utilizing the available low-cost instances provisioned at time \(t\) and sending any remaining traffic to FaaS. 
    When the $\beta$ distribution over early exits drifts, the system triggers targeted re-profiling and re-selection via the $\beta$-Aware Configurator.

    }\label{System_Overview}
\end{figure}

As shown in Fig.~\ref{System_Overview}, \name{} comprises offline and online components.
Given an ingestion rate $\lambda_{t}$ at time $t$ in the online system, the offline system samples a persistent traffic with a long-term average of $N$ req/s and uses the per-instance capacity \(r_{\max}\) to profile VM-only, FaaS-only, and hybrid offloading configurations under various confidence thresholds (\(\mathit{conf\_thres}\)). 
The ingestion rate notations are summarized in Table~\ref{tab:ingestion-rate-notations}.
The offline system comprises:
\begin{itemize}
  \setlength\itemsep{0pt}
  \setlength\parskip{0pt}
  \setlength\parsep{0pt}
  \item \textit{SLO-aware Configurator}, which selects VM and/or FaaS configurations that satisfy the latency SLO.
  \item \proofread{The \textit{$\beta_{pid}$-aware Configurator} evaluates the cost of each candidate configuration based on the distribution of early-exit rates \(\{\beta_{pid}\}\) across different partitions \(pid\).}
\end{itemize}
Using the cost-efficient SLO-satisfying configuration provided by the offline system, the online system consists of the following components:
\begin{itemize}
  \setlength\itemsep{0pt}
  \setlength\parskip{0pt}
  \setlength\parsep{0pt}
  \item \textit{Traffic Monitor}, which updates the EWMA \(\mu\) and moving standard deviation \(\sigma\) of arrival rate \(\lambda_{t}\). 
  It then computes the target VM count \proofread{based on }\(\frac{\mu+\sigma}{r_{\max}}\) and the Traffic Cost Indifference Point (T-CIP), directing the Scaling Manager as needed.
  \item \textit{Scaling Manager}, which provisions or terminates VM instances based on the Traffic Monitor’s instructions.
  \item \textit{Load Balancer}, which routes the stable traffic portion (up to \(\mu+\sigma\) req/s) to VMs (each provisioned at \(r_{\max}\) req/s) and offloads any excess \(\bigl(\lambda_{t}-(\mu+\sigma)\bigr)\) to FaaS.
\end{itemize}

\subsection{Instance Configuration}\label{sec:arch-ofp}
\begin{figure}
\captionsetup[subfigure]{justification=centering}
\centering
    \includegraphics[width=0.48\textwidth]{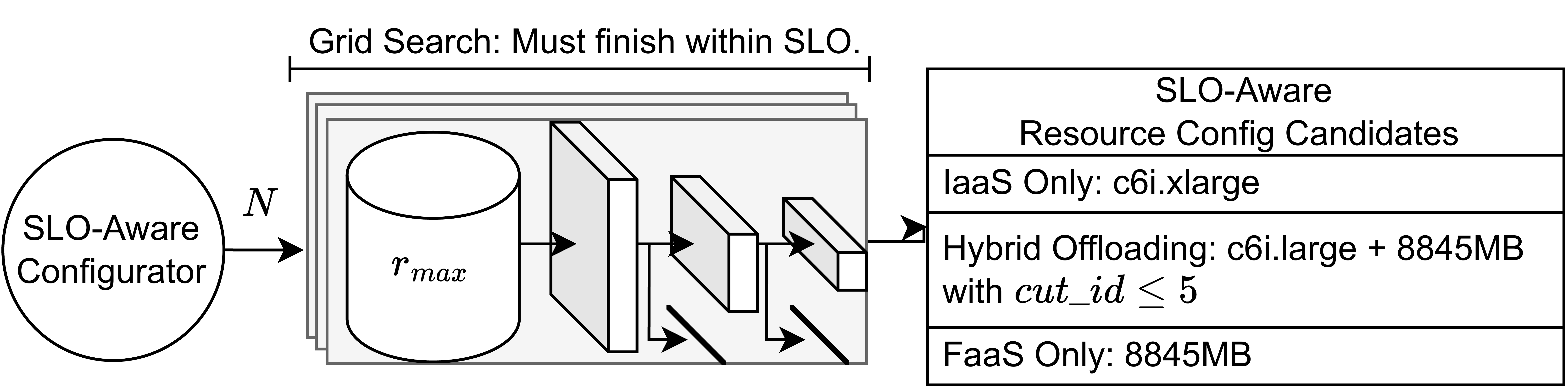}
    \caption[Overview of SLO Aware Configurator]{The SLO-aware Configurator searches for the SLO-compliant configurations given a per-instance ingestion rate $r_{max}$ requests per second for VMs, Hybrid Offloading and Serverless.
    }\label{SLO-Aware-Configurator}
\end{figure}
To minimize cost while meeting latency SLOs online, \name{} first profiles and configures instances offline.
Given a per-instance capacity \(r_{\max}\) (req/s), the \textit{SLO-aware Configurator} enumerates candidate resource configurations, profiles request runtimes, and selects the most cost-efficient configuration whose worst-case latency, when \(r_{\max}\) req/s traverse all model partitions with \(\mathit{conf\_thres}=1\), does not exceed the SLO (Fig.~\ref{SLO-Aware-Configurator}, Alg.~\ref{alg:sla-aware-configurator}).
\begin{figure*}[ht!]
\centering
\captionsetup[subfigure]{justification=centering}
    \subfloat[Hybrid Offloading]{
    \includegraphics[width=0.35\textwidth]{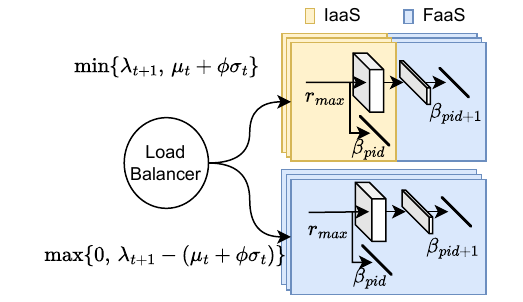}
    \label{fig:Hybrid-Offloading-Setup}}
    \subfloat[IaaS Only]{
    \includegraphics[width=0.35\textwidth]{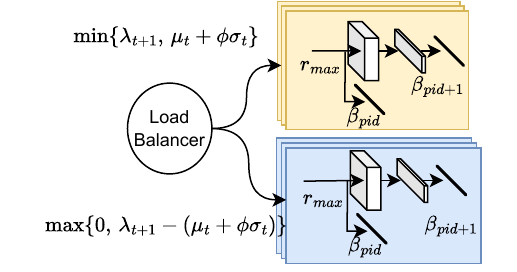}
    \label{fig:IaaS-Setup}}
    \subfloat[FaaS Only]{
    \includegraphics[width=0.25\textwidth]{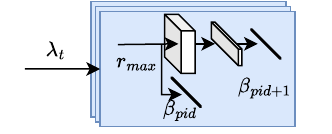}
    \label{fig:FaaS-Setup}}
    \caption[Resource Configuration Setups Selected by Offline System]{Resource configuration setups used by the online system and profiled by the offline system: hybrid offloading (Fig.~\ref{fig:Hybrid-Offloading-Setup}), IaaS only (Fig.~\ref{fig:IaaS-Setup}) and FaaS only (Fig.~\ref{fig:FaaS-Setup}).
    Instances (VM, hybrid offloading, or serverless) are provisioned with capacity \(r_{\max}\) req/s to handle
    stable \(\min\{\lambda_{t+1},\,\mu_{t}+\phi\sigma_{t}\}\) and 
    transient \(\max\{0,\,\lambda_{t+1}-(\mu_{t}+\phi\sigma_{t})\}\)
    traffic with a parameter $\phi \in \mathbb{R}$ from ingestion online traffic $\lambda_{t}$ over time $t$. Notice that we use long-term average of $N$ req/s instead of \(\mu+\sigma\) for profiling the three setups in the offline system.}
    \label{fig:configuration-setups}
\end{figure*}
The \textit{SLO‐aware Configurator} evaluates three candidate setups, VM‐only, FaaS‐only, and Hybrid Offloading, as illustrated in Fig.~\ref{fig:configuration-setups}. 
Here, $N$ denotes the long-term average traffic used for offline profiling and configuration, whereas $\lambda_{t}$ denotes the instantaneous arrival rate observed by the online system at time $t$.

For the FaaS-only setup (Fig.~\ref{fig:FaaS-Setup}), the online system load-balances traffic at a per-function cap \(r_{\max}\) req/s. 
Architects may configure distinct \(r_{\max}\) for IaaS, FaaS, or hybrid offloading to meet latency targets. 
Empirically, for many CPU-intensive FaaS workloads, there is an optimal memory configuration that minimizes the per-invocation cost. This is due to the inherent trade-off in FaaS billing, where higher memory (with a higher per-unit cost) grants more compute and a shorter runtime. While increasing memory size impacts runtime, the cost per invocation often remains within a stable range, though not constant, as compute duration decreases to offset the higher per-unit cost (see Sec.~\ref{sec:eval-offline-profiling}).
On AWS Lambda, allocating multiples of \(1,769\)\,MB avoids hardware sharing.
We thus select \(8,845\)\,MB, the largest such multiple below the \(10,240\)\,MB limit.

\begin{figure}[ht!]
\captionsetup{justification=centering}
 \centering
\includegraphics[width=0.35\textwidth]{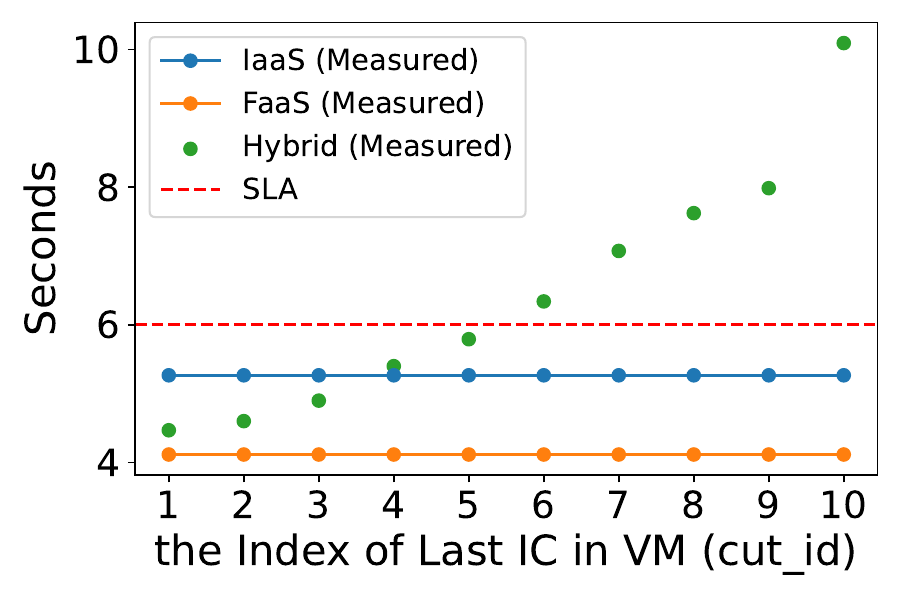}
\caption[Mean Longest Runtime under Different Setups]{
Profiled mean request runtimes, with $r_{\max}=\tfrac{100}{6}$ and \(\mathit{conf\_thres}=1\).
At $cut\_id \leq 5$, the $6$s SLO is maintained. }\label{fig:max_duration}
\end{figure}

For the VM-only and hybrid offloading setups (Figs.~\ref{fig:IaaS-Setup} and~\ref{fig:Hybrid-Offloading-Setup}), the load balancer routes the stable portion of traffic to the VM or hybrid instance and the transient portion to FaaS (Sec.~\ref{sec:arch-olb}). 
Each VM type has distinct cost and capacity.
For our compute‐intensive image classification with \(r_{\max}=100/6\) req/s (100 requests per 6s, SLO = 6s), the configurator selects \texttt{c6i.xlarge} for VM‐only. 
In the hybrid offloading setup, profiling (Fig.~\ref{fig:max_duration}) shows that a \texttt{c6i.large} VM with \(\mathit{cut\_id}\le5\) satisfies the SLO.

\begin{figure}
\captionsetup[subfigure]{justification=centering}
\centering
    \includegraphics[width=0.48\textwidth]{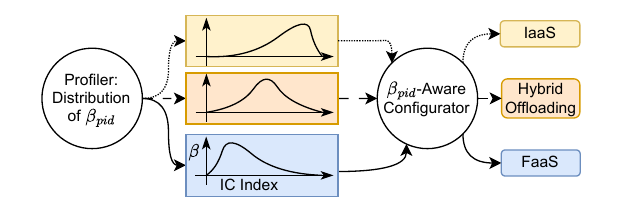}
    \caption[Overview of $\beta_{pid}$ Aware Configurator]{The $\beta_{pid}$-aware Configurator evaluates configuration candidates by profiling the distribution of early exits ($\beta_{pid}$) given an ingestion rate of $N$.}\label{Sparsity-Aware-Configurator}
\end{figure}

Next, the \emph{$\beta_{pid}$-aware Configurator} lowers \(\mathit{conf\_thres}<1\) to enable early exits and measures exit rates \(\{\beta_{pid}\}\) at rate \(N\) req/s.
It then picks the lowest-cost configuration that meets the target accuracy, as shown in Fig.~\ref{Sparsity-Aware-Configurator} and Alg.~\ref{alg:configurator-beta-distribution}.

\begin{figure}[t]
\centering

\begin{minipage}{\linewidth}\centering
  \subfloat[Most exits at shallow ICs.\label{fig:exit_distribution_0.5}]{
    \includegraphics[width=0.45\linewidth]{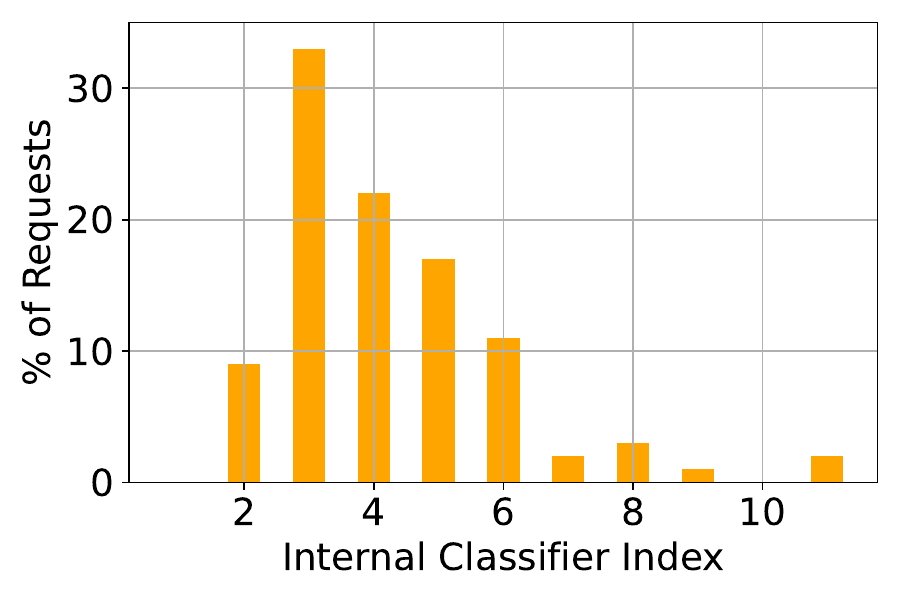}}
  \hfill
  \subfloat[FaaS is cheaper ($cut\_id \leq 5$).\label{fig:0.5_cost_estimated}]{
    \includegraphics[width=0.52\linewidth]{"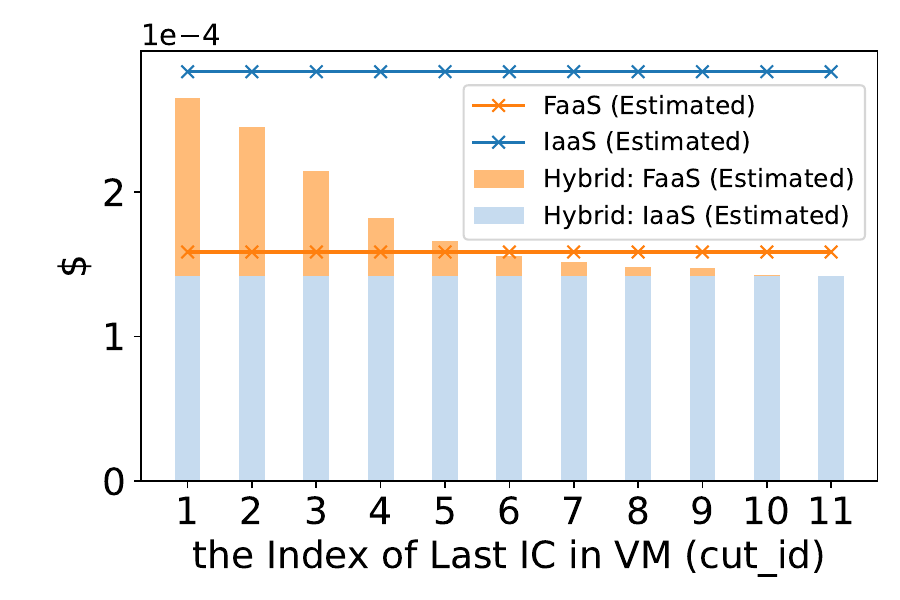"}}
  \par\footnotesize\emph{$conf\_thres = 0.5$}
\end{minipage}


\begin{minipage}{\linewidth}\centering
  \subfloat[Most exits at middle ICs.\label{fig:exit_distribution_0.7}]{
    \includegraphics[width=0.45\linewidth]{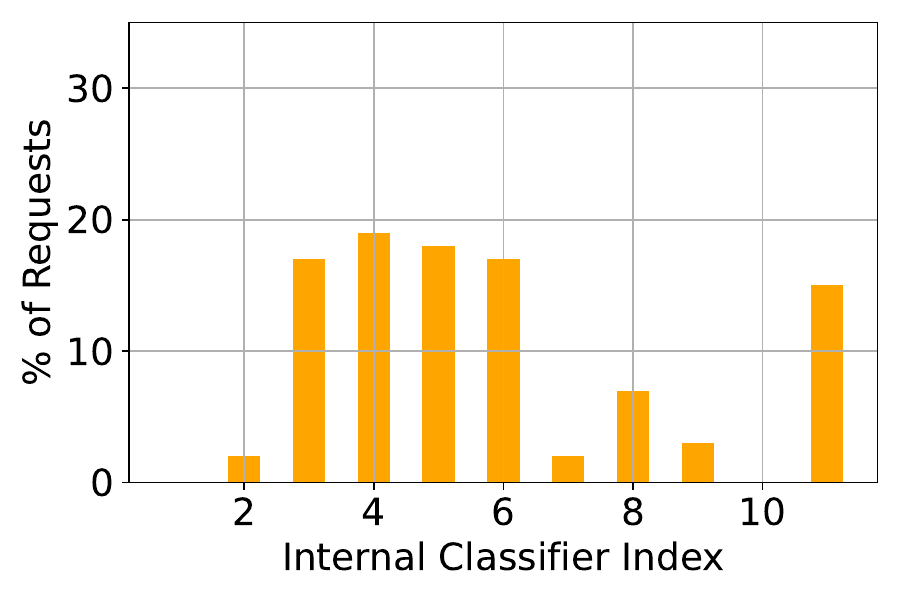}}
  \hfill
  \subfloat[Hybrid is cheaper ($cut\_id \geq 5$).\label{fig:0.7_cost_estimated}]{
    \includegraphics[width=0.52\linewidth]{"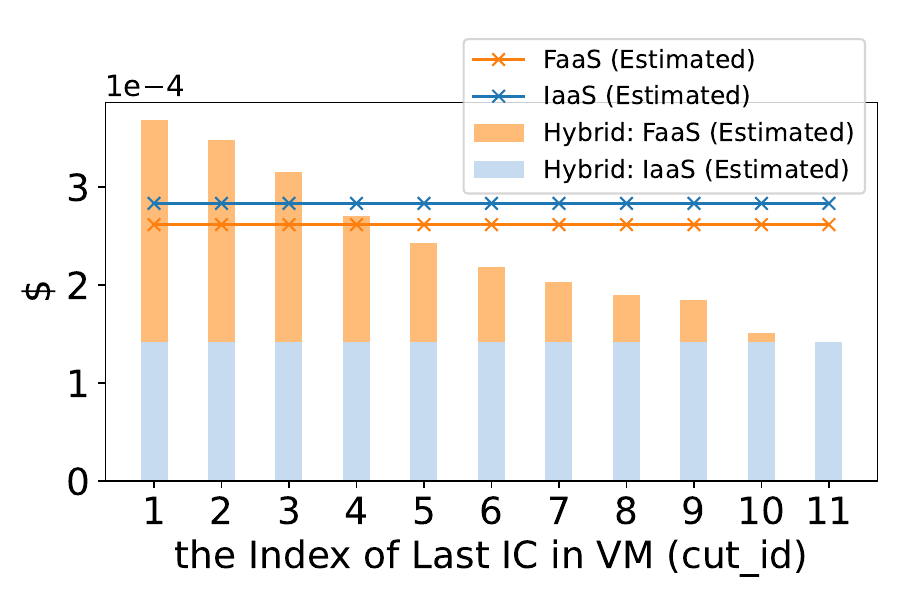"}}
  \par\footnotesize\emph{$conf\_thres = 0.7$}
\end{minipage}


\begin{minipage}{\linewidth}\centering
  \subfloat[Most exits at deep ICs.\label{fig:exit_distribution_0.85}]{
    \includegraphics[width=0.45\linewidth]{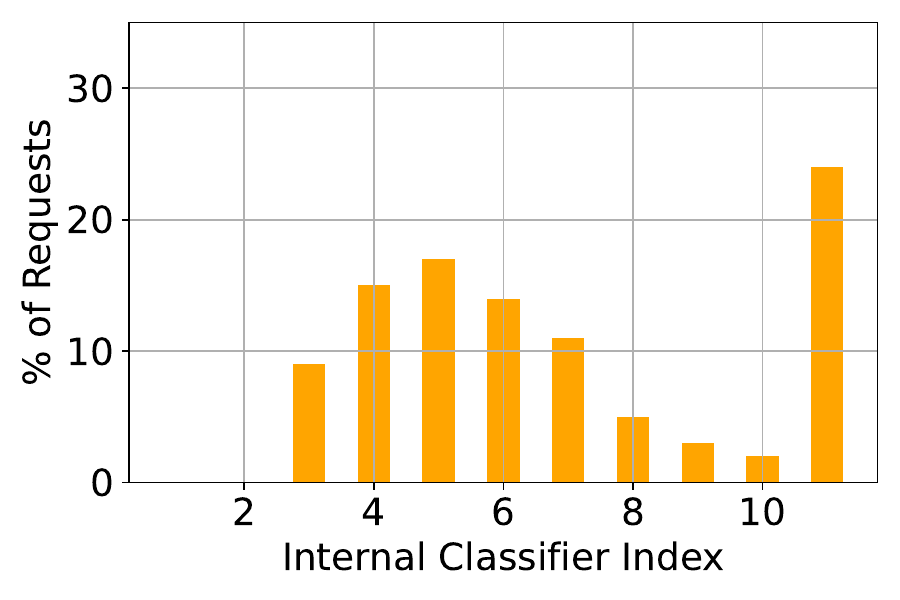}}
  \hfill
  \subfloat[IaaS is cheaper ($cut\_id \leq 5$).\label{fig:0.85_cost_estimated}]{
    \includegraphics[width=0.52\linewidth]{"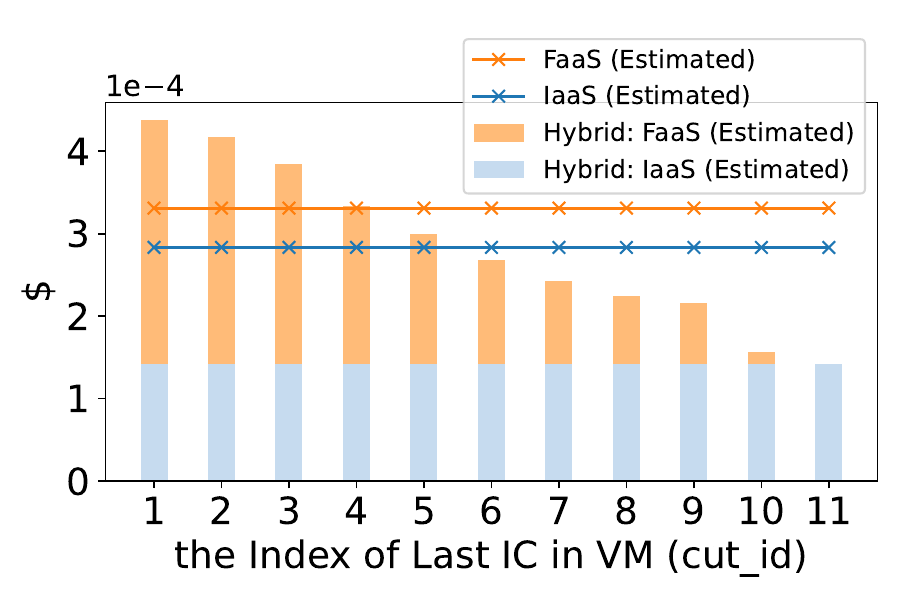"}}
  \par\footnotesize\emph{$conf\_thres = 0.85$}
\end{minipage}

\caption[Cost Comparison under Different $\beta_{pid}$ Distributions]{
Estimated average costs for $N=\frac{100}{6}$ req/s under varying $\beta_{pid}$ distributions and setups. 
At $cut\_id \leq 5$, the SLO holds for all setups (Fig.~\ref{fig:max_duration}), and the cost-minimizing setup shifts with the confidence threshold: FaaS-only at low $conf\_thres$ (e.g., $0.5$), Hybrid Offloading at medium $conf\_thres$ (e.g., $0.7$), and IaaS-only at high $conf\_thres$ (e.g., $0.85$).}
\label{fig:vary-length-eval-estimated}
\end{figure}

We compute the end-to-end cost via Equations~\ref{e2e_cost_iaas_std}, \ref{e2e_cost_faas}, and \ref{e2e_cost_hybrid} and, for each $(N, r_{\max}, \beta_{pid})$ triplet, pick the lowest-cost configuration. 
The $\beta_{pid}$ distributions under $conf\_thres$ (Figs.~\ref{fig:exit_distribution_0.5}, \ref{fig:exit_distribution_0.7}, \ref{fig:exit_distribution_0.85}) yield the estimated costs in Figs.~\ref{fig:0.5_cost_estimated}, \ref{fig:0.7_cost_estimated}, and \ref{fig:0.85_cost_estimated}, which plot monetary cost in USD (y-axis) vs. $cut\_id$\footnote{\proofread{The head portion of the DNN consists of the shallow partitions up to \(\texttt{cut\_id}\), which corresponds to the internal classifier's index.}} (x-axis). 
Homogeneous setups (IaaS-only, FaaS-only) have a fixed cost irrespective of $cut\_id$. 
In hybrid offloading (Fig.~\ref{fig:0.7_cost_estimated}), the VM cost is flat (as VMs are allocated for the full SLO duration), but the FaaS share of the cost drops with increasing $cut\_id$ as more work occurs on the VMs, giving a minimum at $cut\_id=5$ within the SLO.
Increasing $cut\_id$ further breaks the SLO budget (Fig.~\ref{fig:max_duration}).
We summarize the optimal strategy for each $\beta$ distribution:
\begin{itemize}
  \setlength\itemsep{0pt}
  \setlength\parskip{0pt}
  \setlength\parsep{0pt}
  \item \textbf{Early exits (Fig.~\ref{fig:exit_distribution_0.5}).} When most requests exit at shallow internal classifiers, FaaS-only is most cost-efficient (\(C^{F}\le C^{I}\) and \(C^{F}\le C^{H}\) for \(cut\_id\le 5\)) due to under-utilization across VM sizes.
  \item \textbf{Late exits (Fig.~\ref{fig:exit_distribution_0.85}).} When most requests exit at deep internal classifiers, VM-only is optimal (\(C^{I}\le C^{F}\) and \(C^{I}\le C^{H}\) for \(cut\_id\le 5\)), due to lower variance in per-request runtimes.
  \item \textbf{Intermediate exits (Fig.~\ref{fig:exit_distribution_0.7}).} When exits concentrate around middle internal classifiers, hybrid offloading is optimal (\(C^{H}\le C^{F}\) and \(C^{H}\le C^{I}\) at \(cut\_id=5\) while meeting the latency SLO).
\end{itemize}

\subsection{Instance Scaling}
\begin{figure}[ht!]
\centering
    \includegraphics[width=0.5\textwidth]{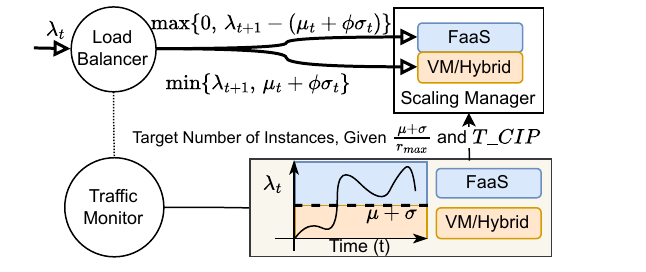}
    \caption[Overview of the Online System]{The online system load-balances $\lambda_{t}$ requests per second and provisions the resources configured by the offline system.}\label{Online-Load-Balancing}
\end{figure}
With the cost-efficient instance configuration, \name{} selects a scaling strategy based on the moving average ingestion rate and the Traffic Cost Indifference Point (T\text{-}CIP).
Fig.~\ref{Online-Load-Balancing} summarizes the online system behaviors.
The Traffic Monitor observes the arrival $\lambda_{t+1}$ and separates persistent and transient load using an exponentially weighted moving average (EWMA) and deviation, updated at time $t{+}1$:
\begin{equation}
    \mu_{t+1} = (1 - W^{\mu})\,\mu_{t} + W^{\mu}\,\lambda_{t+1},
\label{moving_ave}
\end{equation}
\begin{equation}
    \sigma_{t+1} = (1 - W^{\sigma})\,\sigma_{t} + W^{\sigma}\,\bigl|\lambda_{t+1} - \mu_{t+1}\bigr|.
\label{moving_std}
\end{equation}
Accordingly, 
the system scales long-lived instances to the target $\mu_{t+1}+\phi\,\sigma_{t+1}$ (with tunable $\phi\in\mathbb{R}$) for stability.
We set \(\phi=1\) in evaluation. Related work~\cite{LIBRA} studies cost trade-offs when tuning \(\phi\).

Based on the offline $\beta_{pid}$-aware Configurator’s recommendation, 
the Traffic Monitor and Scaling Manager use one of three candidate pools
\[
  \{\text{IaaS},\,\text{FaaS}\},\quad
  \{\text{Hybrid Offloading},\,\text{FaaS}\},\quad
  \{\text{FaaS}\}.
\]
Let $r_{\max}$ denote per-instance service rate (req/s) and define
\[
\begin{aligned}
    k_{t+1}&=\left\lfloor\frac{\mu_{t+1}+\sigma_{t+1}}{r_{\max}}\right\rfloor,\\
    r^{\mathrm{res}}_{t+1}&=(\mu_{t+1}+\sigma_{t+1})-k_{t+1}\,r_{\max}\in[0,\,r_{\max}).
\end{aligned}
\]
We define the \emph{Traffic Cost Indifference Point} (T\text{-}CIP) as a threshold on $r^{\mathrm{res}}_{t+1}$. The Scaling Manager provisions
\[
\#\text{instances}_{t+1}=k_{t+1}+\mathbf{1}\!\left[r^{\mathrm{res}}_{t+1}>\text{T\text{-}CIP}\right],
\]
where the instances are \textit{IaaS} or \textit{Hybrid Offloading} according to the selected pool (no VM is created in the \{\text{FaaS}\} pool). 
Instance scaling depends on the smoothed target $\mu_{t+1}+\sigma_{t+1}$ rather than instantaneous $\lambda_{t+1}$, as illustrated in Fig.~\ref{Online-Load-Balancing} and Alg.~\ref{alg:scalevms}. 
Correspondingly, the online load balancer uses all provisioned IaaS/Hybrid Offloading instances and sends any excess demand (requests) to FaaS, as detailed in Sec.~\ref{sec:arch-olb}.

\begin{figure}[ht!]
\centering
    \includegraphics[width=0.4\textwidth]{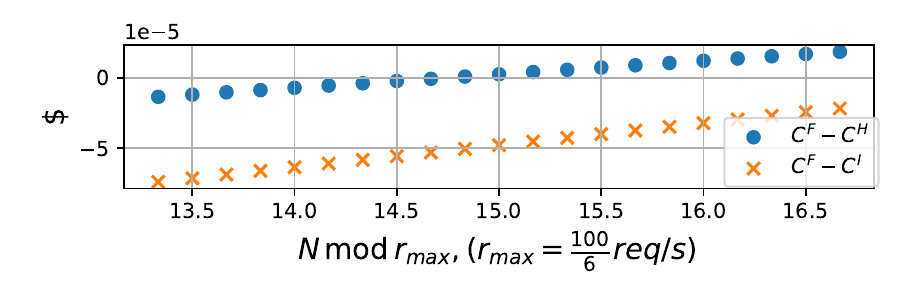}
    \caption[Cost Differences using IaaS, FaaS or Hybrid Offloading Given Ingestion Rate ($N \bmod r_{\max}$)]{
    Cost differences $C^{F}-C^{H}$ and $C^{F}-C^{I}$ as a function of total ingestion rate $N \bmod r_{\max}$, for $r_{\max}=100/6$, c6i.large (Hybrid), c6i.xlarge (IaaS), FaaS at 8845\,MB, a VGG-16 with internal classifiers, and CIFAR-10.
    In Fig.~\ref{fig:F_and_H_and_I_0.7} with $conf\_thres=0.7$, T-CIP between FaaS and Hybrid is $15$ req/s.
    }
    \label{fig:t-cip}\label{fig:F_and_H_and_I_0.7}
\end{figure}

To identify T\text{-}CIP for each pool, the offline configurator uses our cost models to estimate each configuration’s cost as a function of the residual load $N \bmod r_{\max}$. For example, Fig.~\ref{fig:t-cip} plots $C^{F}-C^{H}$ and $C^{F}-C^{I}$ versus $N$ for $r_{\max}=100$, \texttt{c6i.large} (hybrid offloading) or \texttt{c6i.xlarge} (VM-only), $cut\_id=5$, and $conf\_thres=0.7$.
In Fig.~\ref{fig:F_and_H_and_I_0.7}, the point where $C^{F}=C^{H}$ (Hybrid and FaaS have equal cost) occurs at $N \bmod r_{\max}=15$ req/s. For $N \bmod r_{\max}>15$, provisioning a hybrid offloading instance is cheaper; for $N \bmod r_{\max}<15$ req/s, FaaS is more economical.
Meanwhile, there is no point where $C^{F}=C^{I}$. Larger $N$ improves IaaS-only utilization, but at $N \bmod r_{\max}=0$, Hybrid Offloading is still cheaper.\footnote{
We omit the results with $conf\_thres$ favoring FaaS-only or IaaS-only setups (lower or higher $conf\_thres$, respectively) as the profiling steps are the same.}
Thus, in this example, T-CIP = $15$ req/s for Hybrid Offloading instances.
When $r^{\mathrm{res}}_{t+1} \leq 15$, the scaling manager keeps $k_{t+1}$ instances.
Otherwise, it scales to $k_{t+1}+1$ instances.

\begin{algorithm}[ht]
\caption{SLO-Aware Configurator}\label{alg:sla-aware-configurator}
\textbf{Input:}
\begin{itemize}[noitemsep]
  \item $\{N\}$: persistent traffic with long-term average $N$ req/s
  \item $r_{\max}$: req/s per VM
  \item $SLO$: max runtime per request
  \item $\{L_{pid}\}$: DNN layers
  \item $\{\theta_I\}, \{\theta_F\}$: IaaS and FaaS configs
\end{itemize}
\textbf{Output:} cost-optimal $(\theta_I^*,\theta_F^*)$ meeting the SLO

\begin{algorithmic}[1]
  \STATE $S \gets \textsc{Sampling}(\{N\}, \text{count}=r_{\max})$
  \STATE
    $C \gets \{(\theta_I,\theta_F)\mid
      \sum\limits_{L\in L_{pid}}
        T(L,S,\theta_I,\theta_F)
      \le SLO\}$
  \IF{$C\neq\emptyset$}
    \STATE $(\theta_I^*,\theta_F^*) \gets \arg\min_{(\theta_I,\theta_F)\in C}\mathrm{Cost}(\theta_I,\theta_F)$
  \ELSE
    \STATE \textbf{output} “No configuration meets SLO.”
  \ENDIF
\end{algorithmic}
\end{algorithm}

\begin{algorithm}[ht!]
\caption{Configurator: $\beta_{pid}$ Distribution}\label{alg:configurator-beta-distribution}
\textbf{Input:}
\begin{itemize}[noitemsep]
  \item $r_{\max}$: requests/sec per VM
  \item $\{\beta_{pid}\}$: exit-rate distribution
  \item $\{L_{pid}\}$: DNN layers
  \item $\theta_I^0,\ \theta_F^0$: base IaaS/FaaS configs (meet SLO)
  \item $\theta_I^{cand}$: candidate IaaS configs for hybrid setup
  \item $\mathtt{cut\_id}$: offloading cuts
  \item $SLO$: max runtime per request
  \item cost models $C^I(\cdot),\;C^F(\cdot),\;C^H(\cdot)$
\end{itemize}
\textbf{Output:} Optimal \((\theta_I^*,\theta_F^*,\mathrm{cut}_H)\)

\begin{algorithmic}[1]
\STATE $cost_F \gets C^{F}(\theta_{F}^{0})$ \hfill // Cost using only FaaS
\STATE $cost_I \gets C^{I}(\theta_{I}^{0})$ \hfill // Cost using only IaaS
\STATE $cost_H, cut_H \gets \infty, \infty$
\STATE $\theta_{I}^{H} \gets \text{NULL}$
\FORALL{$\theta \in \{\theta_{I}^{cand}\}$}
    \FORALL{$cut \in \{cut\_id\}$}
        \STATE $cost_{H}^{est} \gets C^{H}(\theta,\theta_{F}^{0},cut,\{\beta_{pid}\},\{L_{pid}\},r_{max})$
        \IF{$cost_{H}^{est} < cost_H$}
            \STATE $cost_H, \theta_{I}^{H}, cut_H \gets cost_{H}^{est}, \theta, cut$
        \ENDIF
    \ENDFOR
\ENDFOR
\IF{$cost_I \le \min(cost_F, cost_H)$}
    \STATE $\theta_{I}^* \gets \theta_{I}^{0}$,\; \(\theta_{F}^* \gets \text{NULL}\),\; \(cut_H \gets \text{NULL}\)
\ELSIF{$cost_F \le \min(cost_I, cost_H)$}
    \STATE \(\theta_{I}^* \gets \text{NULL}\),\; \(\theta_{F}^* \gets \theta_{F}^{0}\),\; \(cut_H \gets \text{NULL}\)
\ELSE
    \STATE \(\theta_{I}^* \gets \theta_{I}^{H}\),\; \(\theta_{F}^* \gets \theta_{F}^{0}\)
\ENDIF
\end{algorithmic}
\end{algorithm}

\subsection{Load Balancing}\label{sec:arch-olb}
With a cost-efficient scaling policy, \name{} keeps enough VM/Hybrid Offloading instances to handle mean traffic cheaply.
It must also absorb spikes without waiting for auto scaling.
Fig.~\ref{Online-Load-Balancing} and Alg.~\ref{alg:online-lb} summarize the spike-handling pipeline. 
\name{} sets a load-balancer threshold at \(\mu_{t}+\phi\,\sigma_{t}\) (tunable \(\phi\in\mathbb{R}\)). 
In epoch \(t{+}1\), traffic up to this threshold goes to the long-lived path (IaaS or Hybrid Offloading), and any excess goes to FaaS:
\[
\begin{aligned}
\text{long-lived load} &= \min\{\lambda_{t+1},\,\mu_{t}+\phi\sigma_{t}\},\\
\text{FaaS spill}      &= \max\{0,\,\lambda_{t+1}-(\mu_{t}+\phi\sigma_{t})\}.
\end{aligned}
\]
Because resource provisioning has nonzero latency, the balancer in epoch $t{+}1$ uses instances scaled in epoch $t$. Newly requested capacity becomes available in the next scaling epoch. In our implementation, the balancer fills all available IaaS/Hybrid Offloading instances with the long-lived portion of the demand and forwards the remaining requests to FaaS.
We batch up to $r_{\max}$ per instance. Each batch is routed to an available instance. If none is free, a FaaS function is invoked. Residual requests that do not form a full batch are treated identically: they go to a IaaS/Hybrid Offloading instance when available, otherwise to FaaS (Alg.~\ref{alg:scalevms}).
\name{} depends on accurate profiles of $\{\beta_{pid}\}$ and the ingestion rate $\lambda_{t}$. In practice, a background process would periodically update these profiles and adjust the resource pool. We leave dynamic re-profiling to future work.

\begin{algorithm}
\caption{Online Load Balancer}\label{alg:online-lb}
\textbf{Input:}
\begin{itemize}[noitemsep]
  \item $\lambda_{t}$: req/s, observed at time $t$
  \item $r_{\max}$: requests/sec per instance
\end{itemize}
\begin{algorithmic}[1]
    \STATE $\mu \gets 0$ \hfill\# EWMA mean of requests per second
    \STATE $\sigma \gets 0$ \hfill\# EWMA std.\ dev.
    \FOR{\textbf{each timestamp $t$}} \label{alg:online-lb:start-loop}
        \STATE $images \gets \textsc{DataLoader}(\lambda)$
        \STATE $\mu \gets (1-a)\,\mu + a\,\lambda_{t}$
        \STATE $\sigma \gets (1-b)\,\sigma + b\,|\mu-\lambda_{t}|$
        \STATE $vmThres \gets \mu + \phi \cdot \sigma$
        \STATE $VMCnt \gets \textsc{get\_healthy\_vm}()$
        \STATE \textbf{Concurrently process mini‑batches:}
        \STATE $\textsc{IaaS}(\textsc{Split}(images,\;step=r_{\max})[:VMCnt]$)
        \STATE $\textsc{FaaS}(\textsc{Split}(images,\;step=r_{\max})[VMCnt:]$)
        \IF{$time() \bmod \texttt{scaleInterval} = 0$}
            \STATE \textsc{ScaleVMs}($vmThres$)
        \ENDIF
    \ENDFOR
\end{algorithmic}
\end{algorithm}

\begin{algorithm}[ht!]
\caption{ScaleVMs Function}\label{alg:scalevms}
\textbf{Input:}\\
\quad vmRequests: Number of VM requests\\
\textbf{Output:}\\
\quad requiredVMs: Number of VMs
\begin{algorithmic}[1]
        \STATE $requiredVMs \gets \lfloor vmRequests / r\_max \rfloor$
        \IF{$vmRequests \mod r\_max > T\_CIP$}
            \STATE $requiredVMs \gets requiredVMs + 1$
        \ENDIF
        \STATE \textit{Update Auto Scaling Group to } $requiredVMs$
\end{algorithmic}
\end{algorithm}

\section{Evaluation}\label{sec:eval}

When submodels are independent, \name{} reduces to a classical load-balancing problem.
Each request follows a fixed model sequence with deterministic runtime. 
In contrast, when submodels are dependent and support early exits, \name{} performs stage-aware resource provisioning by accounting for the exit distribution \(\{\beta_{pid}\}\), thereby minimizing end-to-end cloud cost under strict SLOs.
In this evaluation, we focus on the dependent submodel scenario.


\subsection{Experimental Implementation and Setup}
We evaluate \name{} through both offline profiling and online load balancing. Our experiments use a VGG-16 network with internal classifiers on the CIFAR-10 classification task~\cite{cifar10}. 
\zz{Inference latency varies across requests exiting at different classifiers.}

For offline profiling, we measure each VGG-16 partition’s runtime on AWS EC2 (\texttt{c6i.large/xlarge}) and on AWS Lambda across memory configs. 
We also record EC2–Lambda transmission latency in the same AWS region via a public gateway. 
The profiling allows us to estimate the worst-case latency of running the full model and the expected runtime and cost under exit distributions \(\{\beta_{pid}\}\).

For AWS Lambda (Fig.~\ref{fig:fix-flops-eval}), we measure cost at \(\mathit{conf\_thres}=0.85\) across memory sizes and cut indices (\(\mathit{cut\_id}\)). 
The monetary cost is stable whenever memory is an integer multiple of \(1769\,\mathrm{MB}\), ensuring full vCPU allocation. 
Accordingly, we choose \(8845\,\mathrm{MB}\) (5 vCPUs) to minimize resource contention and latency.

\zz{For online evaluation, we replay a sequence of numbers of requests from the WITS traffic trace~\cite{wits_trace} for 400 epochs.}
We batch requests into 6-second SLO intervals in groups of 100 (\(r_{\max}=100/6\) req/s). 
Full batches are sent to available VMs.
Any leftover (including partial) batches are offloaded to Lambda. 
We set the Traffic Cost Indifference Point \(T_{\mathrm{CIP}}=90\), and every 25 epochs (150~s) we scale the VM count based on \(\mu+\sigma\).

VMs are managed by an AWS Auto Scaling Group (disabled auto-scaling) and an Application Load Balancer.
Serverless functions scale automatically. 
In hybrid offloading, EC2 instances send LZ4-compressed hidden variables via POST to Lambda when offloading late-exit requests, making transmission overhead negligible. 
VMs await Lambda responses before returning final logits to clients.

\zz{The online system (Load Balancer, Traffic Monitor, and Scaling Manager) executes on a node in Chameleon Cloud~\cite{chameleoncloud}, polling EC2 and Lambda metrics captured during function calls. }
The implementation of these modules is in Python and can be integrated with ML inference frameworks such as Ray Serve~\cite{ray-serve} on Kubernetes, treating each VM and serverless function as a Ray actor for elastic, large-scale deployment. 
We reserve production-grade integration for future work.


\subsection{Offline Profiling and $S\_CIP$ Configuration}\label{sec:eval-offline-profiling}
This section presents the offline profiling results.
We first identify resource configurations that satisfy the 6-second SLO when all requests traverse every stage.  
\zz{As we discussed in the motivation example (Fig.~\ref{fig:max_duration} in Sec.~\ref{sec:arch-ofp}), for \(r_{\max}=\tfrac{100}{6}\) (100-requests mini-batches every 6 s), IaaS-only setup with a \texttt{c6i.xlarge} under \(\mathit{conf\_thres}=1\) completes each batch in 5.5 s, which is within the SLO of 6 seconds.
In addition, we found that using a \texttt{c6i.large} takes 10.3 s, which would violate the SLO.}
FaaS-only setup using AWS Lambda with \(8845\,\mathrm{MB}\) also meets the SLO in 4.3 s.  
A hybrid offloading setup that runs partitions 1–\(\mathit{cut\_id}\) (with \(\mathit{cut\_id}\le5\)) on \texttt{c6i.large} and offloads the remainder to Lambda with \(8845\,\mathrm{MB}\) also satisfies the 6-second SLO.

\begin{figure}[ht!]
\captionsetup{justification=centering}
 \centering
 \subfloat[Profiled FaaS Cost using Different Memory with $conf\_thres = 0.85$]{\includegraphics[width=0.24\textwidth]{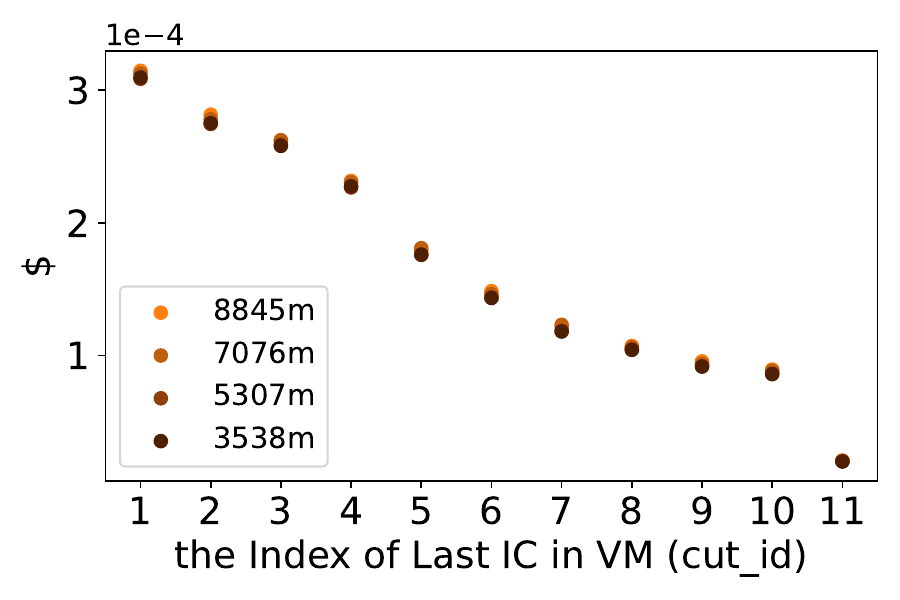}\label{fig:fix-flops-eval}}
 \subfloat[Profiled Costs with $conf\_thres=0.5$]{\includegraphics[width=0.24\textwidth]{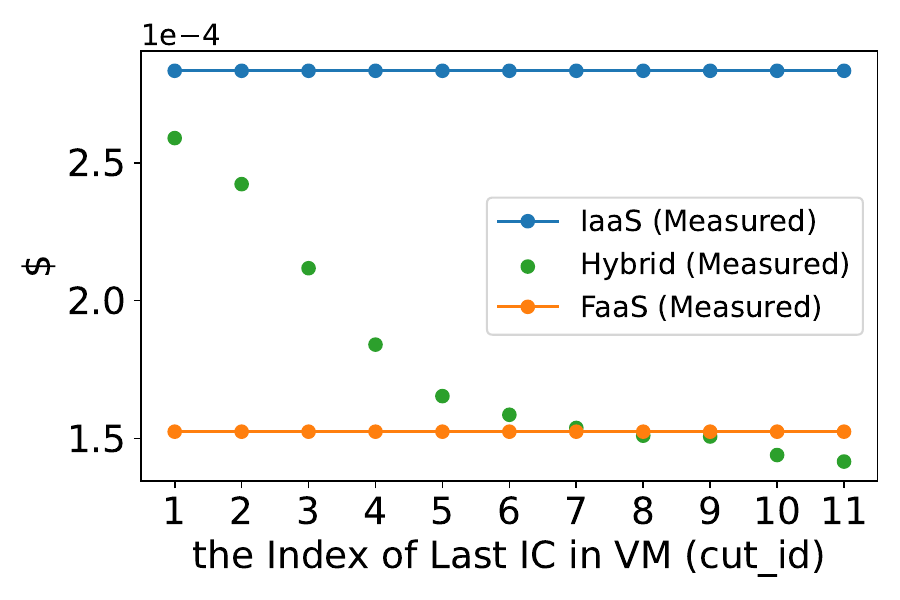}\label{fig:0.5_cost}}\\
  \subfloat[Profiled Costs with $conf\_thres=0.7$]{\includegraphics[width=0.24\textwidth]{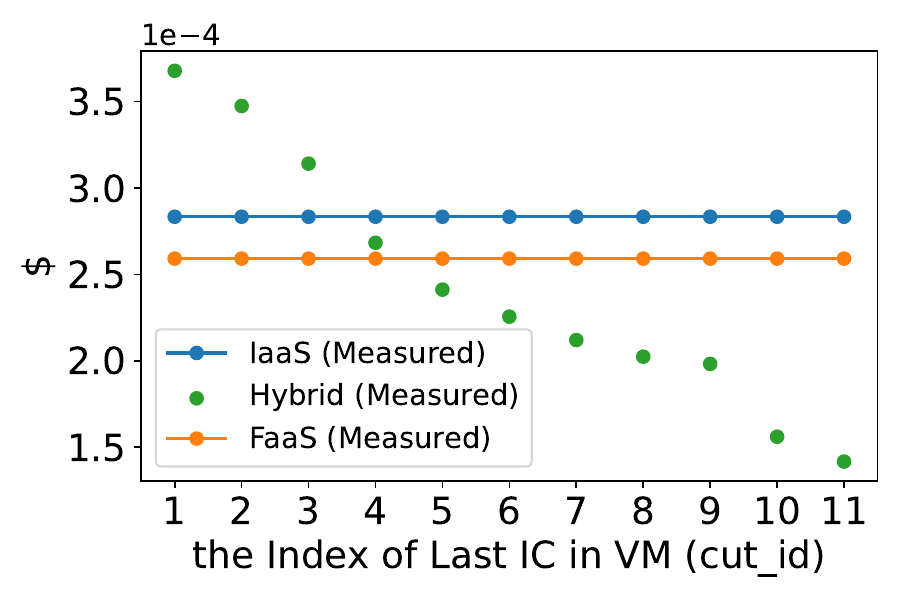}\label{fig:0.7_cost}}
   \subfloat[Profiled Costs with $conf\_thres=0.85$]{\includegraphics[width=0.24\textwidth]{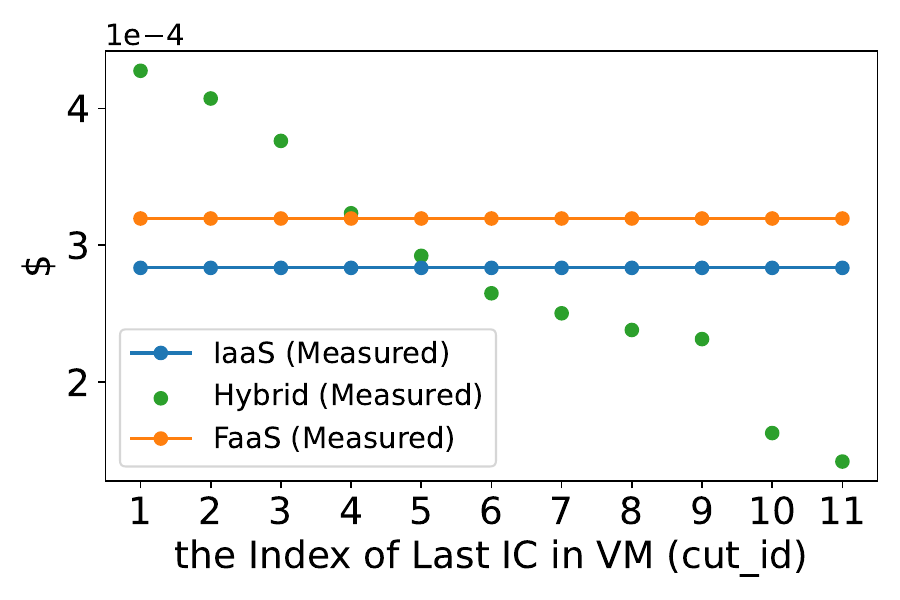}\label{fig:0.85_cost}}
\caption[Profiled Costs at Various $conf\_thres$]{
Fig.~\ref{fig:fix-flops-eval} shows memory has little influence on FaaS resource costs.
Thus, we set $8845$ MB for all test cases.
In Fig.~\ref{fig:0.5_cost}, \ref{fig:0.7_cost} and \ref{fig:0.85_cost}, we show the costs under different $conf\_thres$ of $0.5$, $0.7$ and $0.85$, respectively.
IaaS uses \texttt{c6i.xlarge}.
Hybrid Offloading uses \texttt{c6i.large}. 
Lower thresholds shift exits to shallow ICs, making FaaS more cost‐efficient. 
At \(\mathit{conf\_thres}=0.7\) \zz{and \(cut\_id = 5\)}, hybrid beats FaaS by $8\%$ and IaaS by $15\%$.
} \label{fig:vary-length-eval}
\end{figure}

We profile cost and top-1 accuracy across confidence thresholds \(\mathit{conf\_thres}\in\{0.5,0.7,0.85\}\).  
Figures~\ref{fig:0.5_cost}–\ref{fig:0.85_cost} plot cost with accuracies of 80\%, 85\%, and 86\%, respectively.
\zz{For reference, at \(\mathit{conf\_thres}=1.0\), our evaluation shows that accuracy reaches 87\% (plots are not shown), indicating a minimal loss in accuracy.}
VM costs are flat since VMs run for the full SLO duration to avoid scaling overhead, while FaaS cost varies with execution time.

Each \(\mathit{conf\_thres}\) induces a \(\{\beta_{pid}\}\) distribution. 
Lower thresholds increase early exits at shallow ICs, whereas higher thresholds shift exits toward deeper ICs.  
When \(\mathit{conf\_thres}=0.5\), FaaS-only is the cheapest for any \(cut\_id\le5\), so the online system only provisions \{FaaS\}. 
At \(\mathit{conf\_thres}=0.85\), IaaS-only is the cheapest for any \(cut\_id\le5\), so the online system uses IaaS-only setup which provisions \{IaaS, FaaS\}, matching LIBRA’s setting for stable runtimes. 
At \(\mathit{conf\_thres}=0.7\), the hybrid scheme is the cheapest at \(cut\_id=5\), so the pool of candidates is \{Hybrid, FaaS\}.
We note that the Hybrid Offloading setup can only satisfy SLO when \(cut\_id \leq 5\).

In general, increasing \(cut\_id\) (i.e., assigning more partitions to VMs) reduces VM idle time and hybrid cost. 
Selecting the largest \(cut\_id\) that still meets the SLO maximizes the cost efficiency.
Empirical profiles closely match our estimates in Sec.~\ref{sec:global_optimization}, suggesting that profiling provides an accurate estimation for configurations.
In the next section, we evaluate the online cost performance, focusing on \(\mathit{conf\_thres}=0.7\) and comparing \name{} to LIBRA.

\subsection{Online Load Balancing}
We evaluate load balancing with \(\mathit{conf\_thres}=0.7\), comparing (i) FaaS only, (ii) IaaS+FaaS~\cite{LIBRA}, and (iii) Hybrid Offloading+FaaS. 
We replay the first 400 epochs of the WITS CIFAR-10 classification trace~\cite{wits_trace}, sending requests every SLO interval (6\,s).

\begin{figure}[ht!]
    \centering
    \includegraphics[width=1\linewidth]{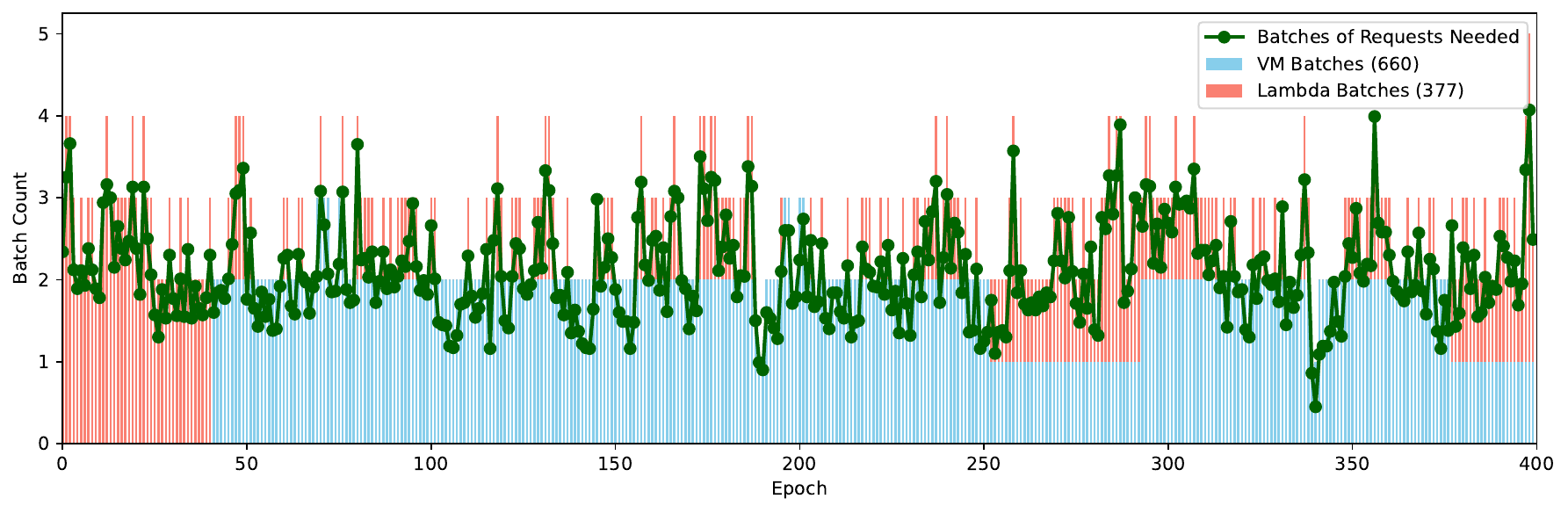}
    \caption[Invocation Batch Type Counts versus Epoch]{The number of mini-batches and how many go to Hybrid Offloading (blue) versus FaaS (red), for \(\mathit{conf\_thres}=0.7\).
    }\label{fig:batch_type_by_epoch}
\end{figure}

\paragraph{Setup:}
Figure~\ref{fig:batch_type_by_epoch} shows, per epoch, the number of mini-batches in total (green line) and handled by FaaS (red bars) versus \zz{Hybrid Offloading (blue bars). }
For example, at epoch~50, there are 336 requests (3.36 mini-batches).
With two healthy VMs, two full batches run on VMs and the remaining one full plus one 36-request batch run on Lambda.

We scale VMs every 25 epochs (150s) using \(\mu + 1\cdot\sigma\), where \(\mu,\sigma\) are the EWMA and moving standard deviation of past load with weights \(W^\mu=W^\sigma=0.5\). 
To avoid overlapping scaling actions, and given VM cold starts may take up to 22 epochs (132\,s), we choose a 25-epoch interval.

\name{} adapts to traffic dynamics.
At epoch 25, it scales to 2 Hybrid Offloading instances, ready by epoch 44.
It scales down to 1 instance at epoch 250 (ready by 252).
And it scales to 2 instances at epoch 275 (ready by 293).

\paragraph{Cost Analysis:}

\begin{figure}[ht!]
    \centering
    \includegraphics[width=0.40\textwidth]{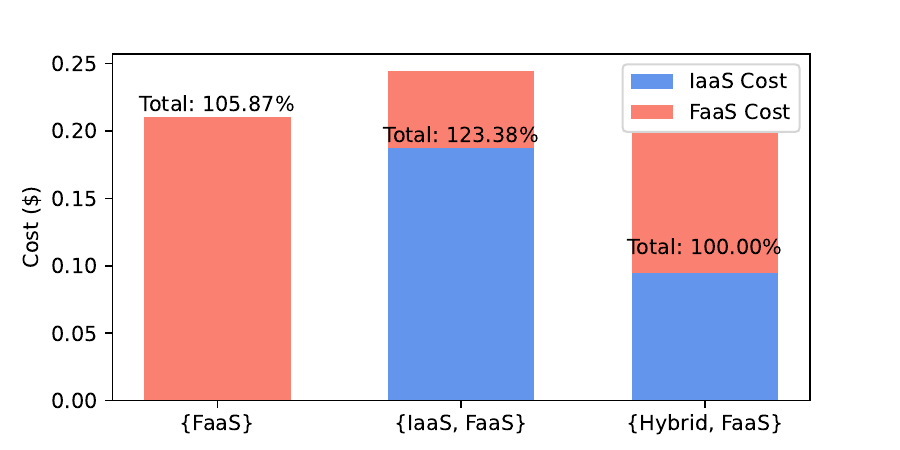}
    \caption[Costs of Using Different Resource Pools with $conf\_thres=0.7$]{Total cost over 400 epochs for \(\{\,\mathrm{FaaS}\}\), \(\{\mathrm{IaaS},\mathrm{FaaS}\}\) (LIBRA), and \(\{\text{Hybrid},\mathrm{FaaS}\}\), with $conf\_thres=0.7$.}
    \label{fig:resource_cost}
\end{figure}

Figure~\ref{fig:resource_cost} compares the total cost of each resource setup over the 400-epoch trace. 
FaaS‐only is $5\%$ more expensive than Hybrid Offloading+FaaS and cannot adapt to sparsely activated models.
And LIBRA’s IaaS+FaaS setup is $20\%$ more expensive than Hybrid Offloading+FaaS. 
All configurations meet the 6s SLO shown by profiling in the prior section.
We evaluate cost with \(\mathit{conf\_thres}=0.7\).
We use \(T\_{CIP}=90\) for Hybrid Offloading (Fig.~\ref{fig:F_and_H_and_I_0.7}), and \(T\_{CIP}=43\) for LIBRA \zz{(when $conf\_thres = 1.0$)}\footnote{\zz{The profiling steps are omitted as they are the same steps as shown in Fig.~\ref{fig:F_and_H_and_I_0.7} with $conf\_thres = 0.7$.}}.

\begin{figure}[ht!]
\captionsetup{justification=centering}
 \centering
  \subfloat[\{Hybrid Offloading, FaaS\}]{\includegraphics[width=0.45\textwidth]{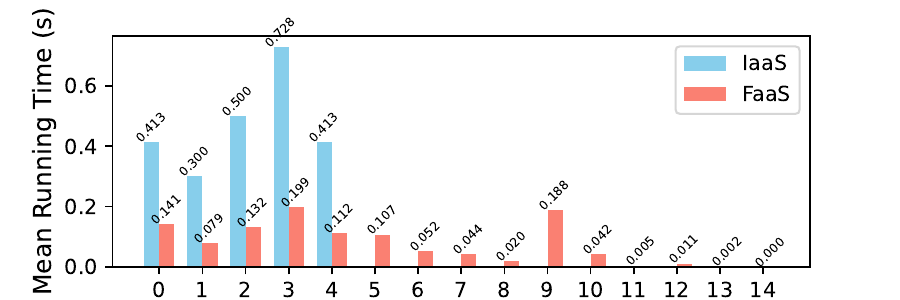}\label{fig:hybrid_pool}}\\
 \subfloat[\{FaaS\}]{\includegraphics[width=0.45\textwidth]{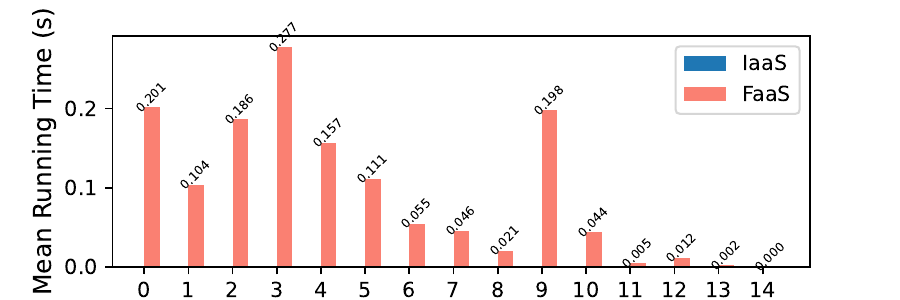}\label{fig:faas_pool}}\\
  \subfloat[\{IaaS, FaaS\}]{\includegraphics[width=0.45\textwidth]{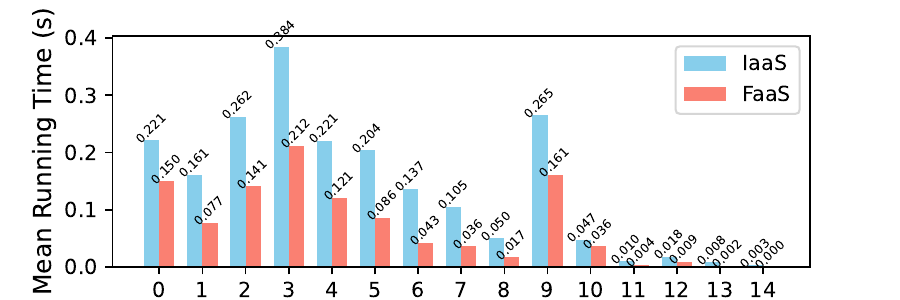}\label{fig:iaas_pool}}
\caption[Average Batch Running Time per Layer]{
Average batch runtime per layer under each resource pool (\(\mathit{conf\_thres}=0.7\)).
} \label{fig:layer_time_0.7}
\end{figure}

\zz{To show VM under-utilization, we plot mean per-layer runtime for each setup in Fig.~\ref{fig:layer_time_0.7}. }
As deeper layers see fewer requests and shorter runtimes, VMs sized for full-model throughput waste capacity.
Thus, in Hybrid Offloading (Fig.~\ref{fig:hybrid_pool}), we place partitions up to \(cut\_id=5\) onto VMs and offload the rest to FaaS. 
This stage-aware matching minimizes idle VM time and lowers overall cost, despite higher per-unit compute cost for deep partitions on FaaS.


\section{Conclusion and Future Work}
\label{sec:conc}

\name{} provides an end-to-end optimization that allocates resources across multiple inference stages in deep neural networks. 
Rather than provisioning for the longest inference chain, it leverages serverless functions for transient workloads and virtual machines for stages with stable demand. 
This hybrid approach reduces cloud costs by over 20\% compared to prior work that assumes uniform runtime.

We envision integrating \name{} into next-generation machine learning-as-a-service platforms, where it acts as a broker to provision cloud services (e.g., AWS Lambda and AWS EC2). 
The online system modules (Load Balancer, Traffic Monitor and Scaling Manager) can be wrapped in ML inference systems like Ray Serve~\cite{ray-serve} on Kubernetes to enable elastic large-scale deployment.

As for future work, \name{} can be extended for other machine learning workloads.
In this work, the resource provisioning is based on the placements of internal classifiers in sequential networks. 
For models with high compute complexity (e.g., large language models), serverless functions may lack sufficient capacity to host even partial models, limiting cost efficiency. 
Future directions include exploring alternative compute platforms with different cost structures or adapting large-scale models for serverless execution.

\bibliography{example_paper}

\begin{thebibliography}{31}
\providecommand{\natexlab}[1]{#1}
\providecommand{\url}[1]{\texttt{#1}}
\expandafter\ifx\csname urlstyle\endcsname\relax
  \providecommand{\doi}[1]{doi: #1}\else
  \providecommand{\doi}{doi: \begingroup \urlstyle{rm}\Url}\fi

\bibitem[amz(2018)]{amzone_auto_scaling}
{Amazon Auto Scaling}.
\newblock \url{https://aws.amazon.com/ec2\\/autoscaling/}, 2018.

\bibitem[goo(2018)]{google_auto_scaling}
{Google Auto Scaling}.
\newblock \url{https://cloud.google.com/compute/\\docs/autoscaler}, 2018.

\bibitem[ama(2020)]{amazon_ec2}
{Amazon EC2}.
\newblock \url{https://aws.amazon.com/ec2/}, 2020.

\bibitem[goo(2020)]{google_compute}
{Google Compute}.
\newblock \url{https://cloud.google.com/compute}, 2020.

\bibitem[{Adobe}(2023)]{adobe-hybridcloud-infra-0}
{Adobe}.
\newblock Cloud infrastructure overview.
\newblock \url{https://experienceleague.adobe.com/en/docs/commerce-operations/implementation-playbook/infrastructure/cloud/overview}, 2023.
\newblock Accessed: 2024-9-11.

\bibitem[{Amazon Web Services}(2024)]{adobe-hybridcloud-infra-1}
{Amazon Web Services}.
\newblock Adobe and aws partnership.
\newblock \url{https://aws.amazon.com/partners/adobe/}, 2024.
\newblock Accessed: 2024-9-11.

\bibitem[{Amazon Web Services}(2025)]{aws_lambda_coldstart}
{Amazon Web Services}.
\newblock {Lambda Runtime Environment: Cold Start Latency}.
\newblock \url{https://docs.aws.amazon.com/lambda/latest/dg/lambda-runtime-environment.html#cold-start-latency}, 2025.
\newblock Accessed: 2025-05-01.

\bibitem[{Anyscale, Inc. and Ray Project contributors}(2025)]{ray-serve}
{Anyscale, Inc. and Ray Project contributors}.
\newblock {Ray Serve: Scalable and Programmable Model Serving}.
\newblock \url{https://docs.ray.io/en/latest/serve/index.html}, 2025.
\newblock Accessed: 2025-07-10.

\bibitem[{AWS}(2024)]{AWS-lambda-mem-config}
{AWS}.
\newblock {Configure Lambda function memory}.
\newblock \url{https://docs.aws.amazon.com/lambda/latest/dg/configuration-memory.html}, 2024.
\newblock Accessed: 2024-10-23.

\bibitem[Behera et~al.(2025)Behera, Champati, Morabito, Tarkoma, and Gross]{behera2025efficient}
Behera, A.~P., Champati, J.~P., Morabito, R., Tarkoma, S., and Gross, J.
\newblock {Towards Efficient Multi-LLM Inference: Characterization and Analysis of LLM Routing and Hierarchical Techniques}.
\newblock \emph{arXiv preprint arXiv:2506.06579}, 2025.
\newblock URL \url{https://arxiv.org/abs/2506.06579}.
\newblock Accessed: 2025-07-10.

\bibitem[{Campbell Webb}(2024)]{workday-hybridcloud-infra}
{Campbell Webb}.
\newblock Unleashing the power of innovation with the public cloud.
\newblock \url{https://blog.workday.com/en-us/unleashing-the-power-innovation-with-public-cloud.html}, 2024.
\newblock Accessed: 2024-9-11.

\bibitem[Castro et~al.(2019)Castro, Ishakian, Muthusamy, and Slominski]{castro2019rise}
Castro, P., Ishakian, V., Muthusamy, V., and Slominski, A.
\newblock {The rise of serverless computing}.
\newblock \emph{Communications of the ACM}, 2019.

\bibitem[Feng et~al.(2025)Feng, Wang, Goyal, Wang, Shi, Xia, Palangi, Zettlemoyer, Tsvetkov, Lee, and Pfister]{heterogeneous-swarms-2025}
Feng, S., Wang, Z., Goyal, P., Wang, Y., Shi, W., Xia, H., Palangi, H., Zettlemoyer, L., Tsvetkov, Y., Lee, C.-Y., and Pfister, T.
\newblock {Heterogeneous Swarms: Jointly Optimizing Model Roles and Weights for Multi-LLM Systems}.
\newblock \emph{arXiv preprint arXiv:2502.04510}, 2025.
\newblock URL \url{https://arxiv.org/abs/2502.04510}.
\newblock Accessed: 2025-07-10.

\bibitem[Gross et~al.(2017)Gross, Ranzato, and Szlam]{gross2017hard}
Gross, S., Ranzato, M., and Szlam, A.
\newblock {Hard Mixtures of Experts for Large Scale Weakly Supervised Vision}.
\newblock In \emph{Proceedings of the IEEE Conference on Computer Vision and Pattern Recognition}, pp.\  6865--6873, 2017.

\bibitem[Gunasekaran et~al.(2019)Gunasekaran, Thinakaran, Kandemir, Urgaonkar, Kesidis, and Das]{spock}
Gunasekaran, J.~R., Thinakaran, P., Kandemir, M.~T., Urgaonkar, B., Kesidis, G., and Das, C.
\newblock {Spock: Exploiting Serverless Functions for SLO and Cost Aware Resource Procurement in Public Cloud}.
\newblock In \emph{2019 IEEE 12th International Conference on Cloud Computing (CLOUD)}, pp.\  199--208, 2019.
\newblock \doi{10.1109/CLOUD.2019.00043}.

\bibitem[Kaya et~al.(2019)Kaya, Hong, and Dumitras]{sdn}
Kaya, Y., Hong, S., and Dumitras, T.
\newblock Shallow-deep networks: Understanding and mitigating network overthinking.
\newblock In \emph{International conference on machine learning}, pp.\  3301--3310. PMLR, 2019.

\bibitem[Keahey et~al.(2020)Keahey, Anderson, Zhen, Riteau, Ruth, Stanzione, Cevik, Colleran, Gunawi, Hammock, Mambretti, Barnes, Halbach, Rocha, and Stubbs]{chameleoncloud}
Keahey, K., Anderson, J., Zhen, Z., Riteau, P., Ruth, P., Stanzione, D., Cevik, M., Colleran, J., Gunawi, H.~S., Hammock, C., Mambretti, J., Barnes, A., Halbach, F., Rocha, A., and Stubbs, J.
\newblock {Lessons Learned from the Chameleon Testbed}.
\newblock In \emph{Proceedings of the 2020 USENIX Annual Technical Conference (USENIX ATC '20)}. USENIX Association, July 2020.

\bibitem[Krizhevsky \& Hinton(2009)Krizhevsky and Hinton]{cifar10}
Krizhevsky, A. and Hinton, G.
\newblock {Learning Multiple Layers of Features from Tiny Images}.
\newblock Technical Report~0, University of Toronto, 2009.
\newblock URL \url{https://www.cs.toronto.edu/~kriz/learning-features-2009-TR.pdf}.

\bibitem[Lewis et~al.(2021)Lewis, Bhosale, Dettmers, Goyal, and Zettlemoyer]{lewis2021base}
Lewis, M., Bhosale, S., Dettmers, T., Goyal, N., and Zettlemoyer, L.
\newblock Base layers: Simplifying training of large, sparse models.
\newblock In \emph{International Conference on Machine Learning}, pp.\  6265--6274. PMLR, 2021.

\bibitem[Little(1961)]{little1961proof}
Little, J.~D.
\newblock A proof for the queuing formula: L= $\lambda$ w.
\newblock \emph{Operations research}, 9\penalty0 (3):\penalty0 383--387, 1961.

\bibitem[Mudvari et~al.(2024)Mudvari, Jiang, and Tassiulas]{splitllm2024}
Mudvari, A., Jiang, Y., and Tassiulas, L.
\newblock {SplitLLM: Collaborative Inference of LLMs for Model Placement and Throughput Optimization}.
\newblock \emph{arXiv preprint arXiv:2410.10759}, 2024.
\newblock URL \url{https://arxiv.org/abs/2410.10759}.
\newblock Accessed: 2025-07-10.

\bibitem[{Novak} et~al.(2019){Novak}, {Kasera}, and {Stutsman}]{feat}
{Novak}, J.~H., {Kasera}, S.~K., and {Stutsman}, R.
\newblock {Cloud Functions for Fast and Robust Resource Auto-Scaling}.
\newblock In \emph{COMSNETS}, 2019.

\bibitem[Purandare et~al.(2023)Purandare, Wasay, and Idreos]{purandare2023mu}
Purandare, S., Wasay, A., and Idreos, S.
\newblock $\mu$-two: 3$\times$ faster multi-model training with orchestration and memory optimization.
\newblock \emph{Proceedings of Machine Learning and Systems}, 5:\penalty0 541--562, 2023.

\bibitem[Raza et~al.(2021)Raza, Zhang, Akhtar, Isahagian, and Matta]{LIBRA}
Raza, A., Zhang, Z., Akhtar, N., Isahagian, V., and Matta, I.
\newblock {LIBRA: An Economical Hybrid Approach for Cloud Applications with Strict SLAs}.
\newblock In \emph{2021 IEEE International Conference on Cloud Engineering (IC2E)}, pp.\  136--146, 2021.
\newblock \doi{10.1109/IC2E52221.2021.00028}.

\bibitem[Shahrad et~al.(2020)Shahrad, Fonseca, Goiri, Chaudhry, Batum, Cooke, Laureano, Tresness, Russinovich, and Bianchini]{serverless-in-the-wild}
Shahrad, M., Fonseca, R., Goiri, I., Chaudhry, G., Batum, P., Cooke, J., Laureano, E., Tresness, C., Russinovich, M., and Bianchini, R.
\newblock {Serverless in the Wild: Characterizing and Optimizing the Serverless Workload at a Large Cloud Provider}.
\newblock In \emph{2020 USENIX Annual Technical Conference (USENIX ATC 20)}, pp.\  205--218. USENIX Association, July 2020.
\newblock ISBN 978-1-939133-14-4.
\newblock URL \url{https://www.usenix.org/conference/atc20/presentation/shahrad}.

\bibitem[Shazeer et~al.(2017)Shazeer, Mirhoseini, Maziarz, Davis, Le, Hinton, and Dean]{shazeer2017}
Shazeer, N., Mirhoseini, A., Maziarz, K., Davis, A., Le, Q., Hinton, G., and Dean, J.
\newblock {Outrageously Large Neural Networks: The Sparsely-Gated Mixture-of-Experts Layer}.
\newblock In \emph{International Conference on Learning Representations}, 2017.
\newblock URL \url{https://openreview.net/forum?id=B1ckMDqlg}.

\bibitem[Shen et~al.(2024)Shen, Lang, Wang, Kim, and Sontag]{shen-etal-2024-learning}
Shen, Z., Lang, H., Wang, B., Kim, Y., and Sontag, D.
\newblock Learning to decode collaboratively with multiple language models.
\newblock In Ku, L.-W., Martins, A., and Srikumar, V. (eds.), \emph{Proceedings of the 62nd Annual Meeting of the Association for Computational Linguistics (Volume 1: Long Papers)}, pp.\  12974--12990, Bangkok, Thailand, August 2024. Association for Computational Linguistics.
\newblock \doi{10.18653/v1/2024.acl-long.701}.
\newblock URL \url{https://aclanthology.org/2024.acl-long.701/}.

\bibitem[Teerapittayanon et~al.(2016)Teerapittayanon, McDanel, and Kung]{BranchyNet}
Teerapittayanon, S., McDanel, B., and Kung, H.
\newblock {BranchyNet: Fast Inference via Early Exiting from Deep Neural Networks}.
\newblock In \emph{2016 23rd International Conference on Pattern Recognition (ICPR)}, pp.\  2464--2469, 2016.
\newblock \doi{10.1109/ICPR.2016.7900006}.

\bibitem[{University of Waikato}(2020)]{wits_trace}
{University of Waikato}.
\newblock {WITS: Waikato Internet Traffic Storage HTTP Dataset}.
\newblock \url{https://wand.net.nz/wits/catalogue.php}, 2020.
\newblock Accessed: 2025-07-10.

\bibitem[Wang et~al.(2018)Wang, Li, Zhang, Ristenpart, and Swift]{peeking-behind}
Wang, L., Li, M., Zhang, Y., Ristenpart, T., and Swift, M.
\newblock {Peeking Behind the Curtains of Serverless Platforms}.
\newblock In \emph{Proceedings of the 2018 USENIX Annual Technical Conference (USENIX ATC)}, Boston, MA, 2018. USENIX Association.

\bibitem[Zhang et~al.(2019)Zhang, Yu, Wang, and Yan]{mark}
Zhang, C., Yu, M., Wang, W., and Yan, F.
\newblock {MArk: Exploiting Cloud Services for Cost-Effective, {SLO}-Aware Machine Learning Inference Serving}.
\newblock In \emph{USENIX ATC}, Renton, WA, 2019.

\end{thebibliography}
\bibliographystyle{mlsys2025}

\end{document}